%% file: neurips2023.tex
\documentclass{article}

\usepackage{xr-hyper}
\usepackage{graphicx}
\usepackage{wrapfig}
\usepackage{tikz}
\usetikzlibrary{bayesnet}
\usepackage{hyperref}
\usepackage{enumitem}
\usepackage{xcolor}
\usepackage{algorithm}
\usepackage{algorithmic}

\newcommand{\blue}[1]{\textcolor{blue}{#1}}
\usepackage{amsmath}
\usepackage{amssymb}
\usepackage{mathtools}
\usepackage{amsthm}
\DeclareMathOperator*{\argmax}{arg\,max}
\DeclareMathOperator*{\argmin}{arg\,min}

\usepackage[nonatbib,final]{neurips2023}

\usepackage{graphicx} 
\usepackage{grffile}

\theoremstyle{plain}
\newtheorem{theorem}{Theorem}[section]
\newtheorem{proposition}[theorem]{Proposition}
\newtheorem{lemma}[theorem]{Lemma}

\theoremstyle{definition}
\newtheorem{definition}[theorem]{Definition}

\theoremstyle{remark}

\usepackage[utf8]{inputenc} 
\usepackage[T1]{fontenc}    
\usepackage{hyperref}       
\usepackage{url}            
\usepackage{booktabs}       
\usepackage{amsfonts}       
\usepackage{nicefrac}       
\usepackage{microtype}      
\usepackage{xcolor}         

\usepackage{makecell}
\title{State Regularized Policy Optimization \\on Data with Dynamics Shift}

\author{%
Zhenghai Xue$^1$ \hspace*{5pt}
Qingpeng Cai$^2$ \hspace*{5pt}
Shuchang Liu$^2$ \hspace*{5pt}
Dong Zheng$^2$ \hspace*{5pt}
\\
\textbf{Peng Jiang}$^2$ \hspace*{5pt}
\textbf{Kun Gai}$^3$ \hspace*{5pt}
\textbf{Bo An}$^1$ \hspace*{5pt}
\\
$^1$Nanyang Technological University, Singapore
\\
$^2$Kuaishou Technology \quad $^3$ Unaffliated
\\
\texttt{zhenghai001@e.ntu.edu.sg}\quad\texttt{boan@ntu.edu.sg}\quad\texttt{gai.kun@qq.com}\\
\texttt{\{caiqingpeng,liushuchang,zhengdong,jiangpeng\}@kuaishou.com}
}

\begin{document}

\maketitle

\begin{abstract}
\input{0.abstract}
\end{abstract}

\section{Introduction}
\input{1.intro}

\section{Backgroud}
\input{2.background}
\section{State Regularized Policy Optimization}
\input{3.srpo}
\section{Theoretical Analysis}
\input{4.theory}

\section{Experiments}
\input{5.experiments}
\section{Conclusion and Discussion}
\input{6.conclusion}

\bibliography{neurips_2023}
\bibliographystyle{unsrt}
\newpage
\input{neurips2023appendix}
\end{document}

%% file: 0.abstract.tex
In many real-world scenarios, Reinforcement Learning~(RL) algorithms are trained on data with dynamics shift, i.e., with different underlying environment dynamics. A majority of current methods address such issue by training context encoders to identify environment parameters. Data with dynamics shift are separated according to their environment parameters to train the corresponding policy.
However, these methods can be sample inefficient as data are used \textit{ad hoc}, and policies trained for one dynamics cannot benefit from data collected in all other environments with different dynamics. 
In this paper, we find that in many environments with similar structures and different dynamics, optimal policies have similar stationary state distributions. We exploit such property and learn the stationary state distribution from data with dynamics shift for efficient data reuse. 
Such distribution is used to regularize the policy trained in a new environment, leading to the SRPO~(\textbf{S}tate \textbf{R}egularized \textbf{P}olicy \textbf{O}ptimization) algorithm. 
To conduct theoretical analyses, the intuition of similar environment structures is characterized by the notion of homomorphous MDPs. We then demonstrate a lower-bound performance guarantee on policies regularized by the stationary state distribution. In practice, SRPO can be an add-on module to context-based algorithms in both online and offline RL settings.
Experimental results show that SRPO can make several context-based algorithms far more data efficient and significantly improve their overall performance.

%% file: 1.intro.tex
Reinforcement Learning~(RL) has achieved great success in solving challenging sequential decision-making problems~\cite{mnih2013playing,silver2016mastering}. Unfortunately, existing RL methods usually assume that agents are trained and evaluated in exactly the same environment, which is often not the case in real-world applications where environment dynamics can vary a lot. For example, the recommendation engine of social apps may need to deal with time-varying and heterogeneous user preferences~\cite{xue2022prefrec,xue2023adarec}.  A robot arm may operate in different scenarios with different joint frictions and medium densities~\cite{liu2022dara}. In these cases, the agent has to work with the trajectory data from \emph{different} environment dynamics, i.e., data with \emph{dynamics shift}, which will bias the learning process and lead to poor performance. In fact, some empirical studies~\cite{luo2022adapt,liu2022dara} demonstrate that general RL algorithms~\cite{schulman2017proximal,haarnoja2018soft} can easily be misled by different environment dynamics and fail to train a good policy. 


In recent years, considerable research efforts have been devoted to addressing the dynamics shift and learning generalizable policies for environments with changing dynamics. One common practice is to train a context encoder~\cite{lee2020context,chen2022latent,zhou2019environment} to associate the environment dynamics with a latent variable. The policy is then trained with the latent variable as an additional input~\cite{chen2021offline}. One issue with this
\begin{wrapfigure}{r}{0.41\textwidth}
    \centering
    \includegraphics[width=0.99\linewidth]{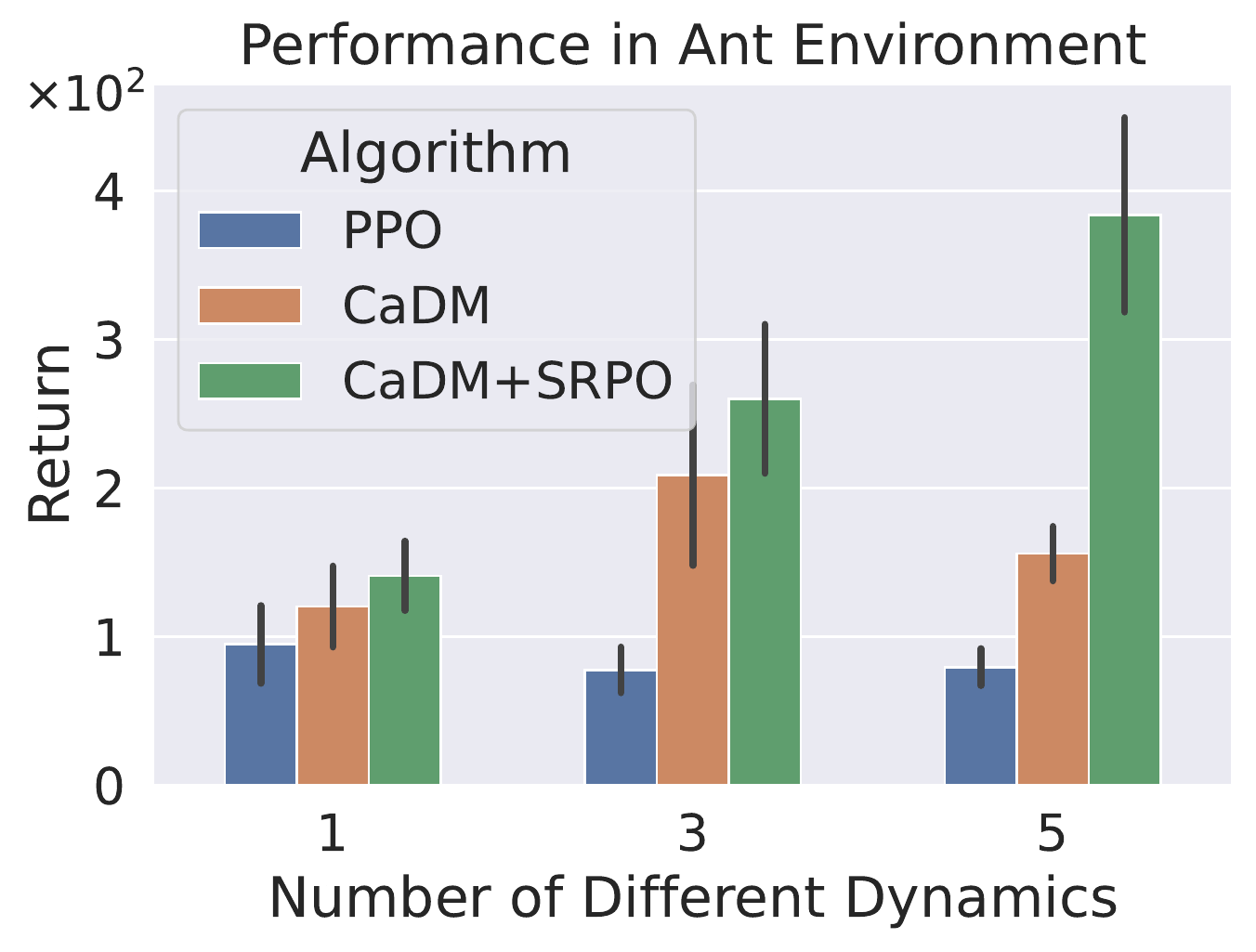}
    \vspace{-0.4cm}
    \caption{Performance comparison of PPO~\cite{schulman2017proximal}, CaDM~\cite{lee2020context} and CaDM+SRPO in the Ant environment, where SRPO is our proposed state regularized policy optimization method. 
    Details of the experiment setup are in Sec.~\ref{sec:exp_setup}.}
    \label{fig:intro}
\end{wrapfigure}
practice is that policies conditioned on a specific latent variable can only learn from data collected in the environment corresponding to that latent variable. In other words, data with different dynamics are used in an \textit{ad hoc} manner. 
The generalizability of context encoders relies on the expressive power of neural networks. However, neural networks are prone to overfit and behave poorly when extrapolating.
As an example, we benchmarked CaDM~\cite{lee2020context}, which is one of the context-based algorithms, under Ant environments with different gravities and display the results in Fig.~\ref{fig:intro}. Although it can outperform PPO~\cite{schulman2017proximal} due to its adaptability from context encoders, CaDM fails to constantly improve its performance with more data from different environment dynamics.
To mitigate the problem of inefficient data use, there are some attempts that leverage Importance Sampling~(IS)~\cite{eysenbach2021off,liu2022dara,niu2022when}.
Given the dynamics of the target environment, samples from the source environments are assigned with larger importance weights if they are more likely to happen in the target environment and vice versa. Compared to training context encoders, IS-based methods manage to proactively \emph{exploit} the data from other dynamics. However, such methods require prior knowledge about the dynamics of the target environment. 
Also, it is notoriously hard to balance the bias and variance when calculating the IS weights.

This paper proposes a new RL paradigm that can explicitly leverage data with dynamics shift. It is also free of the aforementioned drawbacks of IS-based methods. We find that the stationary state distribution induced by optimal policies (later termed optimal state distribution) is similar across a set of environments with similar structures and different environment dynamics.
For example, given heterogeneous preferences of users, a video recommendation system may choose different videos to recommend, but the optimal states are the same: users keep pressing the ``like'' or ``save'' button and continue watching for a long time. 
More concretely, the optimal state distribution in one environment dynamics can be informative for training policies in all other different dynamics. We therefore propose a constrained policy optimization~(CPO)~\cite{joshua2017constrained} formulation that requires the policy not only to optimize the cumulative return, but also to generate a stationary state distribution close to the optimal state distribution. By relating optimality to high-reward states~\cite{levine2018reinforcement}, we are able to approximate the optimal state distribution from trajectory data regardless of the underlying dynamics, providing a unified and efficient approach to exploiting these data.

Summarizing these ideas, we propose the SRPO~(\textbf{S}tate \textbf{R}egularized \textbf{P}olicy \textbf{O}ptimization) algorithm. SRPO works as an add-on module in both online and offline context-based RL algorithms such as CaDM~\cite{lee2020context} and MAPLE~\cite{chen2021offline} to increase their sample efficiency, leading to the CaDM+SRPO and MAPLE+SRPO algorithms. 
We provide a lower-bound performance guarantee on policies in one dynamics regularized by the optimal state distribution in other dynamics. This theoretically demonstrates the effectiveness of the SRPO algorithm in using data with dynamics shift.
Empirical results in both online and offline settings show that SRPO can significantly improve both the data efficiency and the overall performance of several state-of-the-art context-based RL algorithms. 
We also perform ablation studies to demonstrate the effectiveness of each component in the SRPO algorithm.

%% file: 2.background.tex
\subsection{Preliminaries}
\label{sec:pre}
A Markov Decision Process~(MDP) can be defined by a tuple $(\mathcal{S}, \mathcal{A}, T, r, \gamma, \rho_0)$, where $\mathcal{S}$ is the state space, $\mathcal{A}$ is the bounded action space with 
actions $a\in(-1,1)$,
$T(s'|s,a)\in[0,1]$ and $r(s,a,s')\in[-R_{\max},R_{\max}]$ are the transition and reward functions. 
$\gamma\in(0,1)$ is the discount factor and $\rho_0(s)$ is the initial state distribution. In MDPs with deterministic transitions, we denote $T(s,a)$ as the transition function with a slight abuse of notation, and $(T,\varepsilon)$ as $\{T'\mid|T(s,a)-T'(s,a)|<\varepsilon,~\forall s\in\mathcal S, a\in\mathcal A\}$ which is the $\varepsilon-$neighbourhood of $T$. RL aims at maximizing the accumulated return of policy $\pi$: $\eta_T(\pi)$$=E_{\pi,T}\left[\sum\limits_{t=0}^\infty\gamma^t r(s_t,a_t)\right]$, where the expectation is computed with $s_0\sim\rho_0$, $a_t\sim\pi(\cdot|s_t)$, and $s_{t+1}\sim T(\cdot|s_t,a_t)$. The optimal policy $\pi^*$ is defined as $\pi_T^*=\argmax\limits_\pi \eta_T(\pi)$.
In an MDP with a policy $\pi$, the Q-value $Q_T^\pi(s,a)$ denotes the expected return after taking action $a$ at state $s$: $Q_T^\pi(s,a)$$=E_{\pi,T}\left[\sum\limits_{t=0}^\infty\gamma^t r(s_t,a_t)|s_0=s,a_0=a\right]$. The value function is defined as $V_T^\pi(s)=\mathbb E_{a\sim\pi(\cdot|s)}Q_T^\pi(s,a)$ with $V^*_T(s)$ being the shorthand for $V^{\pi_T^*}_T(s)$. It satisfies the optimal Bellman Equation $V^*_T(s)=\max\limits_a~r(s, a)+\gamma\mathbb E_{s'\sim T(\cdot|s,a)} V^*_T(s').$ We can also define the stationary state distribution~(also known as state occupation function) as $d_{T}^\pi(s):=(1-\gamma)$ $\sum_{t=0}^{\infty} \gamma^t P_{T}\left(s_t=s \mid \pi\right)$ with $d^*_T(s)$ being the shorthand for $d^{\pi_T^*}_T(s)$. 

The Hidden Parameter Markov Decision Process~(HiP-MDP) captures a class of MDPs with different transition functions and the same reward function by introducing a set of hidden parameters. Specifically, an HiP-MDP is defined by a tuple $(\mathcal{S}, \mathcal{A}, \Theta, T, r, \gamma, \rho_0)$, where $\Theta$ is the space of hidden parameters. The transition function $T_\theta(s'|s,a,\theta)$ is parameterized not only by states and actions, but also by a hidden parameter $\theta$ sampled from $\Theta$. The action gap of an HiP-MDP is defined as $\Delta=\min\limits_{\theta\in\Theta}\min\limits_{s\in\mathcal{S}}\min\limits_{a\neq\pi^*(s)}V_{T_\theta}^*(s)-Q_{T_\theta}^*(s,a)$, which reflects the minimum gap between an optimal action and all other sub-optimal actions.

\subsection{Related Work}
\label{sec:related}
\paragraph{MDPs with Different Dynamics}
The setting of HiP-MDP~\cite{velez2016hidden} was proposed to model a set of variations in the environment dynamics. The problem is intensively investigated in recent years~\cite{kirk2023survey} and these researches fall into three categories, i.e., encoder-based, Important Sampling (IS)-based and meta-RL based algorithms. Encoder-based methods extract the hidden parameters from trajectories with variational inference~\cite{luisa2020varibad} or auxiliary loss~\cite{lee2020context}.
These hidden parameters are used as inputs to the transition function~\cite{lee2020context} or policy network~\cite{yang2020single,chen2021offline}. 
Unfortunately, these methods train dynamics-specific policies from the trajectory data of each hidden parameter independently, which 
leads to poor sample efficiency. Instead, our method uses the data from all dynamics to learn an optimal state distribution that facilitates the policy learning. IS-based methods compute the importance ratio between transition probabilities under different dynamics and modify the replay buffer~\cite{eysenbach2021off,niu2022when,liu2022dara} according to the transition probabilities in the test environments, which is often not available in real-world scenarios. Finally, meta-RL algorithms~\cite{finn2017model,mitchell2021offline} can adapt to environments with new dynamics through fine-tuning on a small amount of data from the test environment. In contrast, our method can be directly applied to new environment dynamics by making a zero-shot transfer.
\paragraph{Behavior Regularized Methods} 
The idea of constrained policy optimization~(CPO)~\cite{joshua2017constrained} is widely used in RL. Most researches focus on behavior regularized methods, i.e., adding policy constraints based on another policy distribution, as shown in the following optimization problem:
\begin{equation}
\label{eq:related_offline}
    \begin{aligned}
        &\max_\pi~\mathbb E_{s,a\sim \mathcal{D}}\left[\mathbb E_{a'\sim\pi(\cdot|s)}Q(s,a')\right] \qquad\text{ s.t. }~~\mathbb E_{s\sim\mathcal{D}}\left[\hat{D}(\pi(\cdot|s)\|\hat{\pi}(\cdot|s))\right]<\varepsilon,
    \end{aligned}
\end{equation}
where $\mathcal{D}$ is the replay buffer, $\hat{\pi}$ is the regularizing policy, and $\hat{D}$ is a certain distance measure. Maximum-Entropy RL~\cite{tuomas2017reinforcement,haarnoja2018soft} can be considered as CPO with a uniform policy distribution. The sparse action tasks~\cite{pang2021sparsity} can be solved by CPO with a sparse policy distribution. Besides, many offline RL algorithms~\cite{fujimoto2019off,kumar2020conservative,wu2019behavior} are based on the idea of constraining the current policy distribution to be close to the dataset's policy distribution. However, data sampled from environments with different dynamics can have distinct optimal policies, as illustrated in Sec.~\ref{sec:motivating}. In such cases, $\hat{\pi}$ in Eq.~(\ref{eq:related_offline}) may include policies that do not match the current environment, and can therefore be misleading.  So \textit{behavior regularization} in Eq.~(\ref{eq:related_offline}) would fail on data with dynamics shift. Differently, our proposed method is based on \textit{state regularization},
which is more suitable when learning from data with dynamics shift.


\paragraph{Leveraging stationary state distributions}
The stationary state distribution $d^\pi_T(s)$ of policy $\pi$ and dynamics $T$ is an important feature that can measure the differences in policies and transition functions. It has already been exploited in many researches. In Off-Policy RL, Islam et. al~\cite{islam2019off} estimates the stationary state distributions of both the current policy and the mixed buffer policy. It then computes the off-policy policy gradient with the constraint that the two distributions should be close. Some Off-Policy Policy Evaluation (OPE) algorithms~\cite{liu2018breaking,nachum2019dualdice} use the steady-state property of stationary Markov processes to estimate the stationary state distributions. In Imitation Learning~(IL), state-only IL algorithms~\cite{gangwni2020state,liu2020state} requires the stationary state distribution of the current policy to be close to that of the expert policy. In Inverse RL~(IRL), ~\cite{ni2019firl} learns a stationary reward function by computing the gradient of the distance between agent and expert state distribution w.r.t. reward parameters.
In Offline RL, \cite{yang2022regularizing} requires the stationary state distribution of the learning policy and the behavior policy to be close and perform conservative updates. The use of such distributions in our paper is similar to some researches on sim-to-real~\cite{christiano2016transfer,jiang2020offline}. They propose to match the next state distribution in the imperfect simulator and the real environment with inverse dynamics model. They implicitly relies on the idea that the same state distribution should generate similar returns in environments with different dynamics. We formulate the idea in this paper with theorems and quantitatively analyse such similarity in various conditions.

%% file: 3.srpo.tex
\label{sec:3}
In this section, we first give motivating examples on 
why the optimal state distribution in one environment dynamics can be informative in all other different dynamics.
A constrained policy optimization formulation is then proposed in Sec.~\ref{sec:SRPO} based on the optimal state distribution. Solving this optimization problem gives rise to our State Regularized Policy Optimization~(SRPO) algorithm that can leverage data with dynamics shift to improve the policy performance.

\begin{wrapfigure}{r}{0.5\textwidth}
    \centering
    \includegraphics[width=0.98\linewidth]{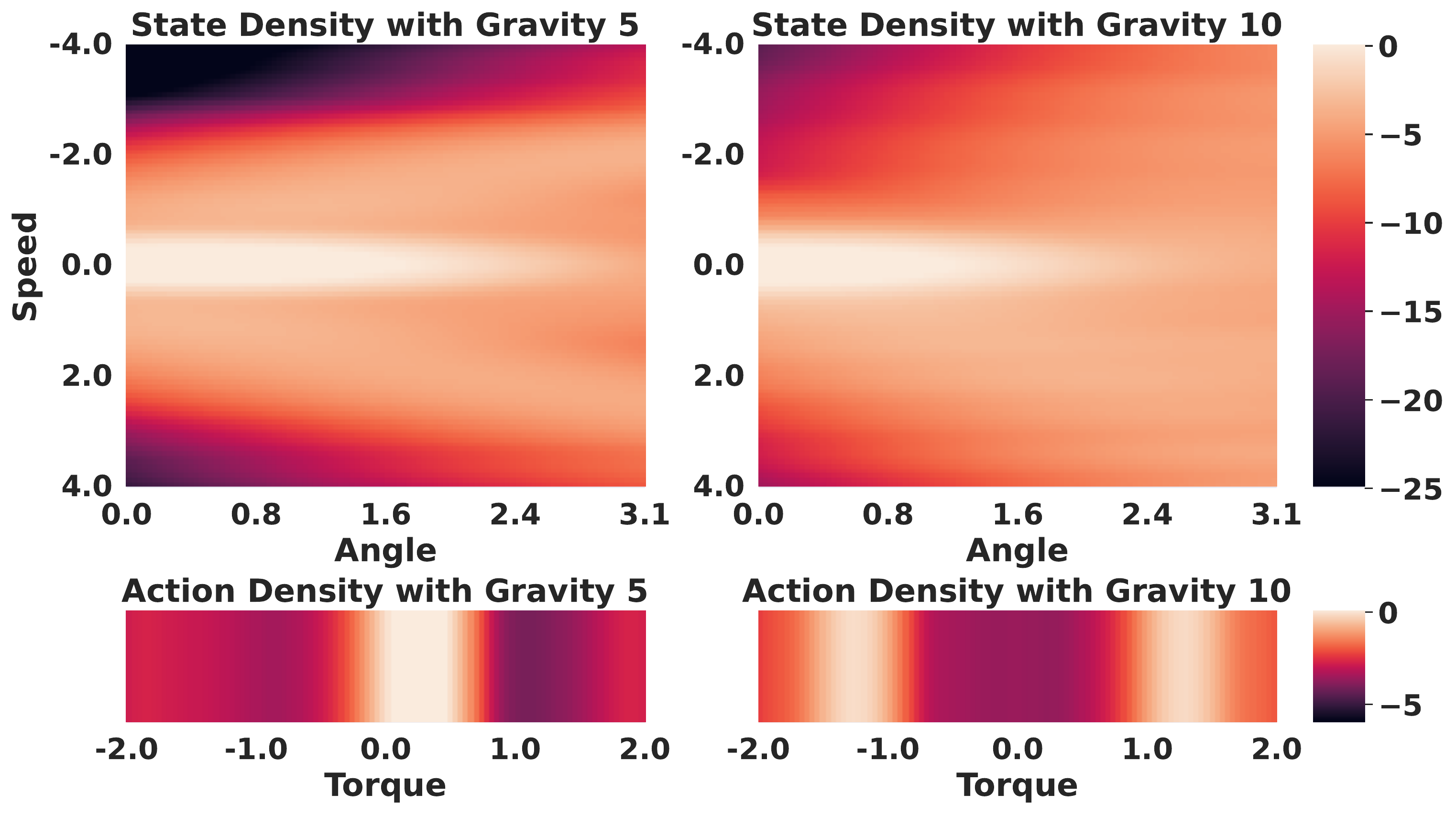}
    \caption{Visualization of state and action densities in data sampled from the Inverted Pendulum environment with gravity 5 and 10. 
    Under both gravities, the state distribution has high density with low pendulum speed and small pendulum angle. Meanwhile, the action distribution has different peaks in density under different gravities.}
    \label{fig:intro_ex}
\end{wrapfigure}
\subsection{Motivating Example}
\label{sec:motivating}
The key intuition behind SRPO is that the optimal state distribution is similar across environments. 
Consider an example of the Inverted Pendulum environment in Fig.~\ref{fig:intro_ex}. We train two policies in the environments with gravities of 5 and 10 until convergence. Then the kernel density estimation~\cite{parzen1962kde} technique is employed to estimate the state and action density of the data collected by the two policies in different areas. It can be observed from the figure that collected data have the same high state density region with low pendulum speed and small pendulum angle, while the action distribution has different density peaks. It demonstrates that the state distribution of data generated by the optimal policy can be similar regardless of the environment dynamics, and therefore can serve as a reference distribution to regularize the training policy in environments with new dynamics. More demonstrating examples can be found at Appendix~\ref{append:exp_analysis}.
 
\subsection{State Regularized Policy Optimization}
\label{sec:SRPO}
Based on the intuition of informative optimal state distribution, we develop a novel technique that regulates RL algorithms to generate a stationary state distribution that is close to the optimal one. Specifically, we propose the following constrained policy optimization formulation:

\begin{equation}
\label{eq:opt_prob}
    \begin{aligned}
    &\max_\pi~\mathbb{E}_{s_t, a_t\sim \tau_\pi} \left[\sum_{t=0}^{\infty} \gamma^t r\left(s_t, a_t\right)\right]\qquad \text{ s.t. }~~D_{\mathrm{KL}}\left(d_\pi(\cdot) \| \zeta(\cdot)\right)<\varepsilon,
    \end{aligned}
\end{equation}
where $\zeta(s)$ is the optimal state distribution in other environment dynamics. 
By introducing the stationary state distribution, the optimization problem defined in Eq.~(\ref{eq:opt_prob}) extends the regularization of in-distribution data to data with distribution shift. A similar form of Eq.~(\ref{eq:opt_prob}) (See Eq.~\ref{eq:related_offline} in Sec.~\ref{sec:related}) is employed in Offline RL algorithms to ensure conservative policy updates. But it restricts the training data to be sampled from the same environment.

We solve Eq.~(\ref{eq:opt_prob}) by casting it to the following unconstrained optimization problem via Lagrange multipliers:
\begin{equation}
\label{eq:res_lag}
    L=-\mathbb E_{s_t,a_t\sim\tau}\left[\sum_{t=0}\limits^\infty\gamma^t\left(r(s_t,a_t)+\lambda\log\frac{\zeta(s_t)}{d_\pi(s_t)}\right)\right]-\frac{\lambda\varepsilon}{1-\gamma},
    \end{equation}
where $\lambda>0$ is the Lagrangian Multiplier. The detailed derivations of the Lagrangian can be found in Appendix~\ref{append:lagrangian}. It is noteworthy that in addition to the multiplier term, the only difference of Eq.~(\ref{eq:res_lag}) and the reward-maximization objective of RL is that the logarithm of probability density ratio $\lambda\log\frac{\zeta(s_t)}{d_\pi(s_t)}$ is added to the reward term $r(s_t,a_t)$. Therefore, one can easily apply our scheme to a wide range of RL algorithms by augmenting the reward function with the density ratio.

\subsection{Data-based Surrogate of the Density Ratio}
\label{sec:surrogate}
The main challenge in solving Eq.~(\ref{eq:res_lag}) is to compute the density ratio $\frac{\zeta(s)}{d_\pi(s)}$ because obtaining the optimal state distribution during online training or given suboptimal offline dataset is infeasible. Also, $d_\pi(s)$ is intractable if the state space is continuous. Motivated by recent advances in adversarial training~\cite{goodfellow2014generative,nowozin2016fgan}, we propose a sample-based surrogate for the density ratio $\frac{\zeta(s)}{d_\pi(s)}$.

\begin{proposition}
    In a GAN, when the real data distribution is $\zeta(s)$ and the generated data distribution is $d_\pi(s)$, the output of the discriminator $D(s)$ follows
    \begin{equation}
        \frac{D(s)}{1-D(s)}=\frac{\zeta(s)}{d_\pi(s)}.
    \end{equation}
\end{proposition}
\begin{wrapfigure}[13]{r}{0.5\textwidth}
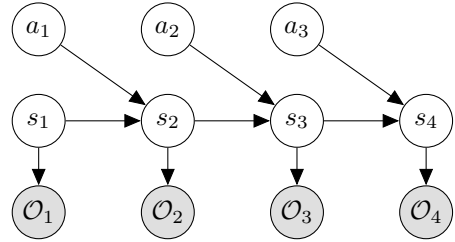

    \centering
    \tikz{
    \node[latent] (s1) {$s_1$};
    \node[latent, above=0.5 of s1] (a1) {$a_1$};
    \node[obs, below=0.5 of s1] (o1) {$\mathcal O_1$};
    \edge{s1}{o1};
    \node[latent, right=of s1] (s2) {$s_2$};
    \node[latent, above=0.5 of s2] (a2) {$a_2$};
    \node[obs, below=0.5 of s2] (o2) {$\mathcal O_2$};
    \edge{a1,s1}{s2};
    \edge{s2}{o2};
    \node[latent, right=of s2] (s3) {$s_3$};
    \node[latent, above=0.5 of s3] (a3) {$a_3$};
    \node[obs, below=0.5 of s3] (o3) {$\mathcal O_3$};
    \edge{a2,s2}{s3};
    \edge{s3}{o3};
    \node[latent, right=of s3] (s4) {$s_4$};
    \node[obs, below=0.5 of s4] (o4) {$\mathcal O_4$};
    \edge{a3,s3}{s4};
    \edge{s4}{o4};
    }
    \caption{HMM in MDP with optimality variables $\mathcal O_t$.}
    \label{fig:hmm}
\end{wrapfigure}
We discuss the relation of this sample-based surrogate with f-divergences and Off-Policy RL in Appendix~\ref{append:fgan}. To train the discriminator $D(s)$, we need to generate samples that is close to the optimal state distribution $\zeta(s)$ and away from $d_\pi(s)$, which is sort of the average state distribution. Motivated by~\cite{levine2018reinforcement}, we model state optimality by a variable $\mathcal O_t$. As shown in Fig.~\ref{fig:hmm}, we regard the state $s_t$ in MDP as a hidden state in a Hidden Markov Model~(HMM), and introduce the binary observation state $\mathcal O_t$. $\mathcal O_t=1$ denotes that $s_t$ is the optimal state at timestep $t$. The observation model is given by
\begin{equation}
    p(\mathcal O_t|s_t)=\max_{a_t}\exp[\gamma^t(r(s_t,a_t)-R_{\max})].
\end{equation}
We can therefore compute the state density ratio $\frac{\zeta(s)}{d_\pi(s)}$ as
\begin{equation}
    \begin{aligned}
        \frac{\zeta(s)}{d_\pi(s)}&=\frac{d_\pi(s|\mathcal{O}_{0:\infty})}{d_\pi(s)}
        =\frac{p(\mathcal O_{0:\infty}|s,\pi)d_\pi(s)}{p(\mathcal O_{0:\infty}|\pi)d_\pi(s)}
        =\frac{\mathbb E_t[p(\mathcal O_{0:t-1}|s_t,\pi)p(\mathcal O_{t:\infty}|s_t,\pi)]}{p(\mathcal O_{0:\infty}|\pi)},
    \end{aligned}
\end{equation}
where the second equation follows the Bayes' law. The last term is related to the forward probability $\alpha_t(s_t)=p(\mathcal O_{0:t-1}|s_t,\pi)$ and backward probability $\beta_t(s_t)=p(\mathcal O_{t:\infty}|s_t,\pi)$ in the HMM. We discuss in Appendix~\ref{append:forward} that $\beta_t(s_t)$ is positively related to a soft version of MDP's state value  $V_\pi(s_t)$. Also, $\alpha_t(s_t)$ makes a little influence on the overall density ratio. Therefore, the input $s$ will be more likely to be sampled from distribution $\zeta(s)$ if it has a higher state value $V(s)$ than average. With this idea, we are able to build training samples for the discriminator $D(s)$. 

\input{algorithm}
\subsection{Practical Algorithm}
\label{sec:prac_algo}
Summarizing the previous derivations, we obtain a practical reward regularization algorithm, termed as SRPO~(\textbf{S}tate \textbf{R}egularized \textbf{P}olicy \textbf{O}ptimization) to leverage data with dynamics shift. We select the MAPLE~\cite{chen2021offline} algorithm, which is one of the SOTA algorithms in context-based Offline RL, as the base algorithm. The detailed procedure of MAPLE+SRPO is shown in Alg.~\ref{alg:main}. After preparing the dataset in a model-based Offline RL style~\cite{yu2020mopo,chen2021offline}, we sample a batch of data from the dataset, obtain a portion of $\rho$ states with higher rewards and add them to the dataset $\mathcal{D}_{\text{real}}$. $\mathcal{D}_{\text{fake}}$ is similarly generated by states with lower rewards (line 10). We set $\rho=0.5$ in offline experiments with medium-expert level of data. $\rho=0.2$ is set in all other experiments. Then a classifier discriminating data from the two datasets is trained (line 11). It estimates the logarithm of the state density ratio $\lambda\log\frac{\zeta(s)}{d_\pi(s)}$, which is added to the reward $r_t$ (line 12). $\lambda$ is regarded as a hyperparameter with values $0.1$ or $0.3$. The effect of $\lambda$ is investigated in Sec.~\ref{sec:exp_analysis}.
The procedure of the online algorithm CaDM~\cite{lee2020context}+SRPO is similar to MAPLE+SRPO, where the datasets $\mathcal{D}_{\text{real}}$ and $\mathcal{D}_{\text{fake}}$ are built with data from the replay buffer, rather than the offline dataset.

%% file: algorithm.tex
\begin{algorithm}[!t]
  \caption{The workflow of SRPO on top of MAPLE~\cite{chen2021offline}.}
  \label{alg:main}
\begin{algorithmic}[1]
  \STATE {\bfseries Input: } $\phi_{\varphi}$ as a context encoder parameterized by $\varphi$; Adaptable policy network $\pi_\theta$ parameterized by $\theta$; Adaptable value network $V_\psi$ parameterized by $\psi$; Offline dataset $\mathcal{D}_{\text {off }}$; Rollout horizon $H$; \blue{State partition ratio $\rho$}; \blue{State discriminator $D_{\delta}$ parameterized by $\delta$}; \blue{Regularization coefficient $\lambda$}.
  \FOR{1, 2, 3, $\dots$}
    \FOR{$t=1$, $2$, $\dots$, $H$}
    \STATE Sample $z_t$ from $\phi_{\varphi}\left(z \mid s_t, a_{t-1}, z_{t-1}\right)$ and then sample $a_t$ from $\pi_\theta\left(a \mid s_t, z_t\right)$.
    \STATE Rollout and get transition data $\left(s_{t+1}, r_{t}, d_{t+1}, s_t, a_t, z_t\right)$. Then add it to $\mathcal{D}_{\text {rollout }}$.
    \ENDFOR
    \STATE Update the context encoder $\phi_{\varphi}$ according to MAPLE.
    \STATE \blue{Sample a batch $\mathcal{D}_{\text{batch}}$ from $\mathcal D_{\text{off}}$ and $\mathcal D_{\text{rollout}}$ and rank them by their state-values estimated by $V_\psi$; Add $\rho|\mathcal{D}_{\text{batch}}|$ states with higher state-values to $\mathcal D_{\text{real}}$ and the others to $\mathcal D_{\text{fake}}$.}
    \STATE \blue{Train the discriminator $D_\delta$ with nll loss.} 
    \STATE \blue{For one-step transition $\left(s_{t+1}, r_{t}, d_{t+1}, s_t, a_t, z_t\right)$ in $\mathcal{D}_{\text{batch}}$, update $r_t$ with $r_t+\lambda\frac{D_\delta(s_t)}{1-D_\delta(s_t)}$}.
\STATE Use the updated $\mathcal D_{\text{batch}}$ and SAC to update the policy and value network parameters $\theta$ and $\psi$.
  \ENDFOR
\end{algorithmic}
\end{algorithm}

%% file: 4.theory.tex
\label{sec:theory}

In this section, we analyze some properties of MDPs with different dynamics and provide theoretical justifications for the SRPO algorithm in Sec.~\ref{sec:3}. The notations are introduced in Sec.~\ref{sec:pre} and proofs can be found in Appendix~\ref{append:proof}.
We first show in Thm.~\ref{th:v_diff2} that the performance of a policy can be lower-bounded when its stationary state distribution is close to a certain optimal state distribution.
In accordance with the intuition in Sec.~\ref{sec:motivating}, it is also demonstrated in Thm.~\ref{cor:a_gap} that optimal policies can induce the same stationary state distribution in different dynamics under mild assumptions.  We start the analysis with the definition of homomorphous MDPs.
\begin{definition}[homomorphous MDPs]
\label{def:homoMDP}
In an HiP-MDP $(\mathcal{S}, \mathcal{A}, \Theta, T, r, \gamma, \rho_0)$, consider hidden parameters $\theta_1,\theta_2\in\Theta$. Let $T_i(s'|s,a)=T(s'|s,a,\theta_i), \forall (s,a,s')\in\mathcal S\times\mathcal A\times\mathcal S,i=1,2$. If $\sum\limits_{a\in\mathcal A} T_1(s'|s,a)>0\Leftrightarrow\sum\limits_{a\in\mathcal A} T_2(s'|s,a)>0$ for all $s,s'\in\mathcal{S}$, MDPs $(\mathcal{S}, \mathcal{A}, T_1, r, \gamma, \rho_0)$ and $(\mathcal{S}, \mathcal{A}, T_2, r, \gamma, \rho_0)$ are referred to as homomorphous MDPs.
\end{definition}
In this definition, $\sum\limits_{a\in\mathcal A} T(s'|s,a)>0$ means state $s'$ can be reached from $s$, so the equivalence of non-zero transition probabilities refers to the same reachability from $s$ to $s'$. Such condition holds in a wide range of MDPs differing only in environment parameters. For example, pendulums with different lengths can all reach the upright state from an off-center state, with longer pendulums exerting a larger force. Apart from the homomorphous property, we also require the reward and dynamics functions of MDPs to have Lipschitz properties. We assume reward function $r(s,a,s')$ w.r.t. the action $a$ is $\lambda_1$-Lipschitz and the dynamics function $T(s,a)$ w.r.t. the action $a$ is $\lambda_2$-inverse Lipschitz. 
%
Discussions on these Lipschitz properties can be found in Appendix~\ref{append:discussions}.

With these preliminaries, we first analyze the discrepancy of accumulated returns of two policies with similar stationary state distributions. The analysis is related to our SRPO algorithm in that the state regularized policy optimization formulation in Eq.~(\ref{eq:opt_prob}) also constrains the learning policy to have a similar stationary state distribution with the optimal policy. Specifically, we derive a theorem as follows.
\begin{theorem}
\label{th:v_diff2}
Consider two homomorphous MDPs with dynamics $T$ and $T'$. If $T'\in(T, \varepsilon_m)$, for all learning policy $\hat{\pi}$ such that $D_{\mathrm{KL}}(d^{\hat{\pi}}_T(\cdot)\|d^{*}_{T'}(\cdot))\leqslant\varepsilon_s$, we have
 \begin{equation}
\label{eq:return_diff}
    \eta_T(\hat{\pi})\geqslant\eta_T(\pi_T^*)-\dfrac{\lambda_1\lambda_2\varepsilon_m+2\lambda_1+\sqrt{2}R_{\max}\sqrt{\varepsilon_s}}{1-\gamma}.
\end{equation}

\end{theorem}
The theorem implies that if a policy $\hat{\pi}$ has a similar stationary state distribution with the optimal policy in one MDP $M$, $\hat{\pi}$ will have a lower-bound performance guarantee in all MDPs that are homomorphous with the MDP $M$. Therefore, the learning policy can benefit from the state regularized policy optimization in Sec.~\ref{sec:SRPO}. 

More specifically, Eq.~(\ref{eq:return_diff}) shows that the gap in accumulated return of $\hat{\pi}$ and $\pi^*_{T}$ is related to the dynamics shift $\varepsilon_m$,  the KL-Divergence of two stationary state distributions $\varepsilon_s$, 
and the effective planning horizon $\frac{1}{1-\gamma}$. With respect to the dynamics shift $\varepsilon_m$, it is related to a ``uniform'' constraint on the dynamics $T'$. We further show in Appendix~\ref{append:proof} that constraining the dynamics shift on a certain state-action pair is enough to derive Eq.~(\ref{eq:return_diff}). Unlike the dynamics shift $\varepsilon_m$ that is determined by a pre-defined RL task, the discrepancy between stationary state distributions $\varepsilon_s$ 
is determined by the learning policy $\hat{\pi}$ and can be optimized during training to obtain a better performance lower-bound.
We also discuss in Appendix~\ref{append:discussions} how tight Eq.~(\ref{eq:return_diff}) is in terms of the effective planning horizon $\frac{1}{1-\gamma}$, compared with some similar performance bounds.

With an additional assumption on the action gap $\Delta$ (defined in Sec.~\ref{sec:pre}), we further demonstrate that the optimal policy of two homomorphous MDPs can have the same stationary state distribution, which verifies the intuition in Sec.~\ref{sec:motivating}.
\begin{theorem}
\label{cor:a_gap}
Consider two homomorphous MDPs with dynamics $T$ and $T'$. If $T'\in(T, \varepsilon_m)$ and the action gap $\Delta$ follows $\Delta>\frac{(2-\gamma)\lambda_1\lambda_2\varepsilon_m}{1-\gamma}$, for all $s\in\mathcal S$ we have $d^{*}_T(s)=d^{*}_{T'}(s)$.
\end{theorem}
\vspace{-0.1cm}
The assumption is mild and holds in many scenarios. For example, in autonomous driving it can be very dangerous to deviate from the optimal policy. Such suboptimal actions have low rewards, leading to a large action gap $\Delta$. In recommendation tasks we are hardly concerned with what items we recommend (the action), as long as the recommendation outcome (the state), i.e., the users' experiences are good enough, leading to a small $\lambda_1$. The condition of large enough action gap holds in these situations. 

%% file: 5.experiments.tex
In this section, we conduct experiments to investigate the following questions: (1) Can SRPO leverage data with distribution shift and outperform current SOTA algorithms in the setting of HiP-MDP, in both online and offline RL? (2) How does each component of SRPO~(e.g., use state regularization rather than behavior regularization) contribute to SRPO's performance? To answer question (1), we use the MuJoCo simulator~\cite{todorov2012mujoco} and generate environments with different transition functions. We train the CaDM+SRPO and the MAPLE+SRPO algorithm proposed in Sec.~\ref{sec:prac_algo} and make comparative analysis with baseline algorithms. To answer question (2), we do ablation studies to examine the role of different modules in SRPO. We also examine how the discriminator $D_\delta$ works in complex environments and the effect of regularizing with state distributions in different performance levels.
\begin{figure}[t]
    \centering
    \includegraphics[width=0.99\linewidth]{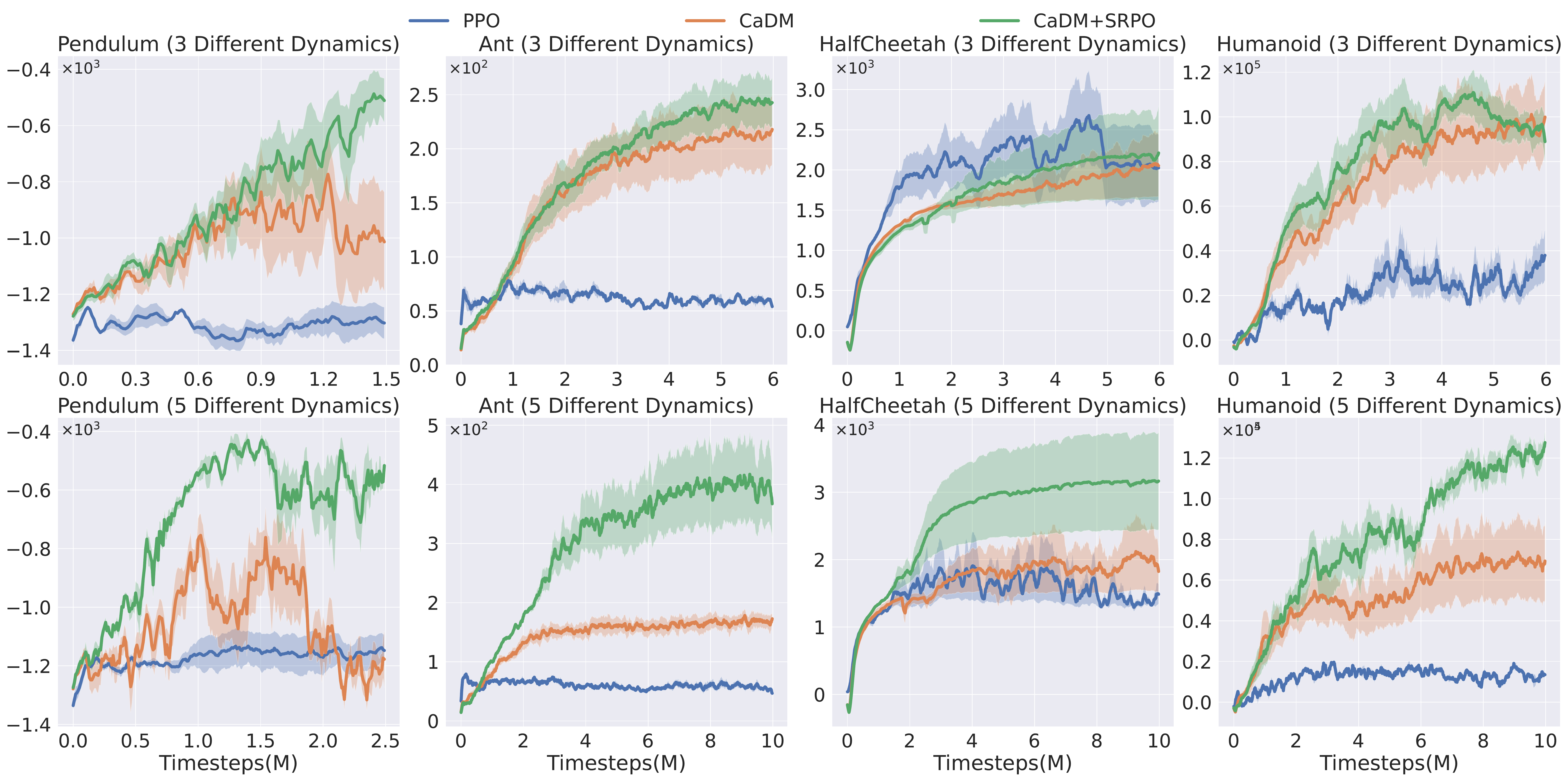}
    \caption{Results of online experiments on MuJoCo tasks. The comparison is made between CaDM+SRPO and baseline algorithms PPO, CaDM. Our CaDM+SRPO algorithm has the best overall performance in experiments with 3 and 5 different environment dynamics. The curves show the average return on 4 random seeds and the shadow areas reflect the standard deviation.}
    \label{fig:online_exp}
\end{figure}
\subsection{Experiment Setup}
\label{sec:exp_setup}
We alter the simulator gravity to generate different dynamics in online experiments. Possible values of gravity are \{1.0\}, \{0.7,1.0,1.3\}, and \{0.4,0.7,1.0,1.3,1.6\} in experiments with 1, 3, and 5 kinds of different dynamics, respectively. When the simulator resets, the gravity is uniformly sampled from the set of all possible values.  The number of training steps is in proportion to the number of environment parameters. Therefore, the agent has access to the same amount of training data on a certain value of simulator gravity. 
We also consider the shift of medium density and body mass in offline experiments to show SRPO's robustness to different forms of dynamics shift. 

To perform comparative analysis, we choose CaDM~\cite{lee2020context} and PPO~\cite{schulman2017proximal} as baseline algorithms in online experiments. In offline experiments, DARA~\cite{liu2022dara} also exploits large amount of data with dynamics shift. Its algorithm relies on Importance Sampling and will be used as a baseline method. Apart from that, we choose MOPO~\cite{yu2020mopo}, MAPLE~\cite{chen2021offline} and CQL~\cite{kumar2020conservative} as baseline methods.  
More information on the setup of experiments is shown in Appendix~\ref{append:exp_setup}.

\input{offline_table.tex}
\subsection{Results}
\label{sec:exp_res}
\paragraph{Online Experiments}
\label{sec:online}
The results of online experiments are shown in Fig.~\ref{fig:online_exp}. With the context encoder and conditional policy, CaDM is able to outperform PPO in all environments. However, it fails to take advantage of the increase in the amount of data with dynamics shift. Its performance with 5 different dynamics is lower than that with 3 dynamics. In contrast, our proposed SRPO algorithm leads to better performance on top of CaDM in accordance with more training data. It significantly outperforms the original CaDM algorithm in environments with 5 different dynamics. The performance comparison in the Pendulum environment is also in accordance with the motivating example in Sec.~\ref{sec:motivating}.
More results of online experiments are shown in Appendix~\ref{append:exp_res}.

\paragraph{Offline Experiments}

The results of offline experiments are shown in Tab.~\ref{tab:offline}. The column of ``CQL Single'' refers to the evaluation score in the CQL~\cite{kumar2020conservative} paper, where the policy is with data from a single static environment. Without the mechanism of context-based encoders, GAIL~\cite{jonothan2016generative}, CQL and MOPO~\cite{yu2020mopo} cannot handle data with distribution shift and show a performance drop.  MAPLE~\cite{chen2021offline} and MAPLE+DARA~\cite{liu2022dara} only achieve marginal performance improvement with respect to CQL single. On the other hand, MAPLE+SRPO shows significant performance improvement over CQL single, which means that SRPO can efficiently leverage the additional data with dynamics shift to facilitate policy training. The MAPLE+SRPO algorithm also has a 15\% higher evaluation score than MAPLE, achieving the best performance in 8 out of 12 tasks. Apart from MAPLE, the meta-RL algorithm PEARL~\cite{kate2019efficient} also has an context encoder for fast adaptation. We compare PEARL with PEARL+SRPO and leave the results in Appendix.~\ref{append:exp_analysis}.

\subsection{Analysis}
\label{sec:exp_analysis}
\paragraph{Ablations}
We conduct ablation studies in offline environments to analyze the role of each algorithm component in SRPO. The results are shown in Tab.~\ref{tab:ablation}. We first investigate the outcome of regularizing with state-action distribution rather than state distribution in Eq.~(\ref{eq:opt_prob}). The resulting policy has a lower evaluation score on average than policies trained with the original SRPO algorithm in all environments. This is because environments with different dynamics do not have a similar optimal policy. The action distribution in the mixed dataset can be misleading when training new policies. According to Sec.~\ref{sec:surrogate}, SRPO trains a classifier to discriminate states with higher values from lower values. We also train another classifier discriminating a random binary partition of states. Ablation results show a huge performance drop, which verifies the effectiveness of the classification-based surrogate mechanism. We also evaluate MAPLE+SRPO with a fixed value of the hyperparameter $\lambda$. $\lambda=0.1$ is more suitable for Walker2d and HalfCheetah environments, while in the Hopper environment $\lambda=0.3$ is better. This is in accordance with previous analysis that Hopper agents can benefit more from regularizing with the stationary state distribution.
\paragraph{Effectiveness of Discriminators} 
\begin{wrapfigure}[14]{r}{0.38\textwidth}
    \centering
    \vspace{-0.3cm}
    \includegraphics[width=0.98\linewidth]{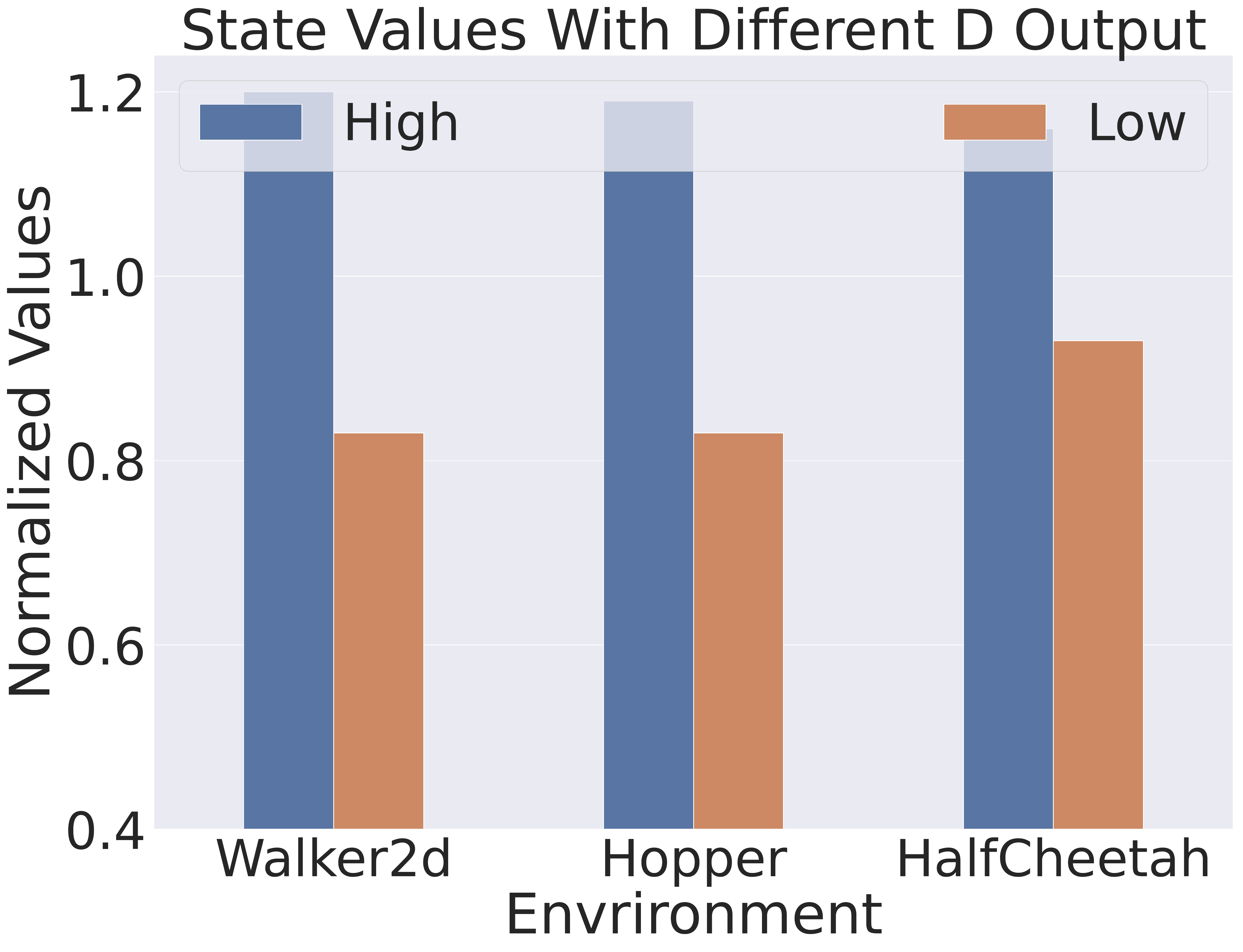}
    \caption{Comparison of values on states with high and low output of the discriminator $D$.}
    \label{fig:analysis}
\end{wrapfigure}
We first train a discriminator $D_\delta$ according to Alg.~\ref{alg:main}. Then a set of states is sampled from the D4RL~\cite{fu2020d4rl} dataset and classified into two sets according to the output of $D_\delta$. The average values of the two sets of states are compared in Fig.~\ref{fig:analysis}. As shown in the figure, states with higher $D_\delta$ outputs also have higher values in all three environments. It means that the trained discriminator $D_\delta$ can successfully identify states with high values from those with low values. Therefore, its output can be a good surrogate for the density ratio $\frac{\zeta(s)}{d_\pi(s)}$ in Sec.~\ref{sec:3}.

\paragraph{Effectiveness of the Regularization}
We also study the effect of policy regularization with different performance levels of stationary state distributions. Random, medium and expert policies in the original Hopper environment are used to estimate the stationary state distributions, which regularize the learning policies in a new environment with different dynamics. The results are shown in Tab.~\ref{tab:analysis}, where the expert policy is the most effective in regularizing. This verifies the practice in Sec.~\ref{sec:SRPO} and the theoretical analysis.

%% file: offline_table.tex
\newcommand{\rowl}[8]{#7 & #8 & #4 & #1 & #2 & #6 & #5 \\}
\begin{table}[t]
\scriptsize
    \centering
    \caption{Results of offline experiments on MuJoCo tasks. Numbers are the normalized scores according to the D4RL paper~\cite{fu2020d4rl}. ME, M, MR and R correspond to the medium-expert, expert, medium-replay and random dataset, respectively. The evaluation is done on policies at the last iteration of training, averaged over four random seeds. The number after $\pm$ is the standard deviation. Our proposed MAPLE+SRPO algorithm has the best performance in 8 of 12 tasks and the highest overall performance.}
    \label{tab:offline}
    \begin{tabular}{l|c|c|c|c|c|c|c|}
    \toprule
        ~ & \rowl{MOPO }{ MAPLE }{ MAPLE-Single }{ CQL }{ \makecell{MAPLE\\+SRPO(Ours)} }{ \makecell{MAPLE\\+DARA} }{ \makecell{CQL\\(Single Env)} }{GAIL}
        \midrule
        Walker2d-ME & \rowl{ 0.25$\pm$0.18 }{ 0.55$\pm$0.21 }{ 0.74 }{ 1.03$\pm$0.10 }{ 0.66$\pm$0.08 }{ 0.80$\pm$0.02 }{ \textbf{1.11} }{0.21$\pm$0.03}
        Walker2d-M & \rowl{ 0.23$\pm$0.34 }{ 0.82$\pm$0.01 }{ 0.56 }{ 0.78$\pm$0.01 }{ \textbf{0.84}$\pm$0.03}{ 0.83$\pm$0.03 }{ 0.79 }{0.15$\pm$0.06}
        Walker2d-MR & \rowl{ 0.00$\pm$0.00 }{ 0.16$\pm$0.02 }{ 0.76 }{ 0.07$\pm$0.00 }{ 0.17$\pm$0.02 }{ 0.17$\pm$0.01 }{ \textbf{0.27} }{0.00$\pm$0.00}
        Walker2d-R & \rowl{ 0.00$\pm$0.00 }{ \textbf{0.22}$\pm$0.00}{0.22 }{ 0.03$\pm$0.01 }{ \textbf{0.22$\pm$0.00 }}{ \textbf{0.22}$\pm$0.00 }{0.07 }{0.00$\pm$0.00}
        Hopper-ME &\rowl{ 0.01$\pm$0.00 }{ 0.96$\pm$0.14 }{ 0.43 }{ 0.32$\pm$0.14 }{ \textbf{0.98}$\pm$0.02 }{ 0.96$\pm$0.06 }{ \textbf{0.98}}{0.04$\pm$0.01 }
        Hopper-M &\rowl{ 0.01$\pm$0.00 }{ 0.78$\pm$0.28 }{ 0.21 }{ 0.57$\pm$0.16 }{ \textbf{1.03}$\pm$0.09 }{ 0.40$\pm$0.05 }{ 0.58}{0.00$\pm$0.00 }
        Hopper-MR &\rowl{ 0.01$\pm$0.01 }{ 0.91$\pm$0.11 }{ 0.88 }{ 0.14$\pm$0.02 }{ \textbf{1.02}$\pm$0.01 }{ \textbf{1.02}$\pm$0.01 }{ 0.46}{0.00$\pm$0.00 }
        Hopper-R & \rowl{0.01$\pm$0.00 }{ 0.13$\pm$0.00 }{ 0.11 }{ 0.11$\pm$0.00 }{ \textbf{0.32}$\pm$0.02 }{ 0.13$\pm$0.01 }{ 0.11 }{0.00$\pm$0.00 }
        HalfCheetah-ME &\rowl{ -0.03$\pm$0.00 }{ 0.50$\pm$0.06 }{ 0.64 }{ 0.03$\pm$0.04 }{ \textbf{0.63}$\pm$0.01 }{ 0.50$\pm$0.00 }{ 0.62}{0.36$\pm$0.06 }
        HalfCheetah-M &\rowl{ 0.38$\pm$0.28 }{ 0.62$\pm$0.01 }{ 0.50 }{ 0.43$\pm$0.03 }{ 0.63$\pm$0.01 }{ \textbf{0.67}$\pm$0.03 }{ 0.44 }{0.25$\pm$0.02}
        HalfCheetah-MR & \rowl{-0.03$\pm$0.00 }{ 0.52$\pm$0.00 }{ 0.59 }{ 0.46$\pm$0.00 }{ \textbf{0.55}$\pm$0.00 }{ 0.53$\pm$0.01 }{ 0.46 }{0.18$\pm$0.11}
        HalfCheetah-R &\rowl{ -0.03$\pm$0.00 }{ 0.22$\pm$0.03 }{ 0.38 }{ 0.01$\pm$0.02 }{ 0.24$\pm$0.01 }{ 0.21$\pm$0.00 }{ \textbf{0.35} }{0.14$\pm$0.02}
        \midrule
        Average &\rowl {0.068 }{ 0.53 }{ 0.50 }{ 0.33 }{ \textbf{0.61} }{ 0.54 }{ 0.52}{0.11 }
        \bottomrule
    \end{tabular}
\end{table}
\newcommand{\rows}[5]{#1 & #2 & #3 & #5\\}
\begin{table}
\footnotesize
\begin{minipage}[t]{0.63\linewidth}
    \centering 
    \caption{Results of ablation studies in offline experiments.}
\vspace{0.05cm}
    \label{tab:ablation}
    \begin{tabular}{l|c|c|c|c|c|c|}
    \toprule
        ~ & \rows{ \makecell{MAPLE\\+SRPO} }{\makecell{Behavior\\Regularizing}}{\makecell{Random\\ Partition}}{\makecell{Fixed \\$\lambda=0.1$}}{\makecell{Fixed \\$\lambda=0.3$}}
        \midrule
        Walker2d& \rows{\textbf{0.47}}{0.45}{0.39}{0.44}{0.40}
        Hopper& \rows{\textbf{0.83}}{0.68}{0.56}{0.67}{0.79}
        HalfCheetah& \rows{\textbf{0.51}}{0.50}{0.45}{0.50}{0.40}
        Average & \rows{\textbf{0.61}}{0.54}{0.47}{0.54}{0.53}
        \bottomrule
    \end{tabular}
\end{minipage}
\begin{minipage}[t]{0.37\linewidth}
\centering
    \caption{Performance comparison of differently regularized policies.}
    \label{tab:analysis}
    \begin{tabular}{|c|c|c|}
    \toprule
        ~ & \makecell{Original\\Hopper Env} & \makecell{10x \\Density} \\
        \midrule
        Random & 121.3 & 44.15 \\
        Medium & 2178 & 913.3 \\
        Expert & 3819 & 3748 \\
        \bottomrule
    \end{tabular}
\end{minipage}
\end{table}

%% file: 6.conclusion.tex
In this work, we focus on the problem of leveraging data with dynamics shift to efficiently train RL agents. Based on the intuition that optimal policies can lead to similar stationary state distributions, we give a constrained optimization formulation that regards the state distribution as a regularizer. After discussions on a sample-based surrogate, we propose the SRPO algorithm which can be an add-on module to context-based algorithms and improve their sample efficiency. The resulting CaDM+SRPO and MAPLE+SRPO algorithms show superior performance when learning on data sampled from environments with different dynamics. Theoretical analyses are also given to analyze some properties of MDPs with different dynamics. They provide justifications for the intuition of the dynamics-invariant state distribution, as well as the constrained policy optimization formulation.

\paragraph{Limitations and Future work} The theoretical analyses of this work requires the assumption of homomorphous MDPs, i.e., the same state reachability in different MDPs. It would be interesting to discuss whether similar conclusions on the stationary state distribution can be derived without such assumption.

\paragraph{Acknowledgements} We thank Wanqi Xue and Yanchen Deng for helpful discussions. This research is supported by the National Research Foundation, Singapore under its Industry Alignment Fund – Pre-positioning (IAF-PP) Funding Initiative and Ministry of Education, Singapore, under its Academic Research Fund Tier 1 (RG13/22). Any opinions, findings and conclusions or recommendations expressed in this material are those of the author(s) and do not reflect the views of National Research Foundation, Singapore.

%% file: neurips2023appendix.tex
\appendix

\section{Additional Derivations and Proofs}
\subsection{Derivations of the Lagrangian}
\label{append:lagrangian}
We start from the optimization problem:
\begin{equation}
    \begin{aligned}
    &\max_\pi~\mathbb{E}_{s_t, a_t\sim \tau_\pi} \sum_{t=0}^{\infty} \gamma^t r\left(s_t, a_t\right)\\
    &\text{ s.t. }~~D_{\mathrm{KL}}\left(d_\pi(\cdot) \| \zeta(\cdot)\right)<\varepsilon_m.
    \end{aligned}
\end{equation}
The KL-Divergence term can be transformed as:
\begin{equation}
\begin{aligned}
    D_{\mathrm{KL}}\left(d_\pi(\cdot) \| \zeta(\cdot)\right)
&=-\mathbb{E}_{s \sim d_\pi(s)}\left[\log \zeta(s)-\log d_\pi(s)\right]\\
&=-\int d_\pi(s)\left[\log \zeta(s)-\log d_\pi(s)\right] d s\\
&=-\int(1-\gamma) \sum_{t=0}^{\infty} \gamma^t p\left(s_t=s\right)\left[\log \zeta(s)-\log d_\pi(s)\right] d s\\
&=-(1-\gamma) \sum_{t=0}^{\infty} \int \gamma^t p\left(s_t=s\right)\left[\log \zeta(s)-\log d_\pi(s)\right] d s\\
&=-(1-\gamma) \sum_{t=0}^{\infty} \mathbb{E}_{s_t \sim \tau}\left[\gamma^t\left(\log \zeta(s_t)-\log d_\pi(s_t)\right)\right]\\
&=-(1-\gamma) \mathbb{E}_{s_t \sim \tau} \sum_{t=0}^{\infty}\gamma^t\left(\log \zeta(s_t)-\log d_\pi(s_t)\right).
\end{aligned}
\end{equation}
So the constraint can be written as
\begin{equation}
    \mathbb{E}_{s_t \sim \tau} \sum_{t=0}^{\infty}\left[\gamma^t\left(\log d_\pi(s_t)-\log \zeta(s_t)\right)\right]-\frac{\varepsilon_m}{1-\gamma}<0.
\end{equation}
The optimization problem can be written as the following standard form
\begin{equation}
    \begin{aligned}
    &\min_\pi~\mathbb{E}_{s_t, a_t\sim \tau} \sum_{t=0}^{\infty} -\gamma^t r\left(s_t, a_t\right)\\
    &\text{  s.t. }~~\mathbb{E}_{s_t \sim \tau} \sum_{t=0}^{\infty}\left[\gamma^t\left(\log d_\pi(s_t)-\log \zeta(s_t)\right)\right]-\frac{\varepsilon_m}{1-\gamma}<0.
    \end{aligned}
\end{equation}
So the Lagrangian $L$ is
\begin{equation}
    L=-\mathbb E_{s_t,a_t\sim\tau}\left[\sum_{t=0}^\infty\gamma^t\left(r(s_t,a_t)+\lambda\log\zeta(s_t)-\lambda \log d_\pi(s_t)\right)\right]-\frac{\lambda\varepsilon_m}{1-\gamma}.
\end{equation}

\subsection{Derivations of the Forward and Backward Probabilities}
The backward probability can be written as:
\label{append:forward}
\begin{equation}
    \begin{aligned}
        \beta_t(s_t)&=\int_{\mathcal S} p(\mathcal O_{t:\infty}|s_t,s_{t+1},\pi)p(s_{t+1}|s_t)ds_{t+1}\\
        &=\int_{\mathcal S} p(\mathcal O_{t}|s_t,\pi)p(\mathcal O_{t+1:\infty}|s_{t+1},\pi)p(s_{t+1}|s_t)ds_{t+1}\\
        &=\int_{\mathcal S} \max_{a_t}\exp(\gamma^tr(s_t,a_t))\beta_{t+1}(s_{t+1})p(s_{t+1}|s_t)ds_{t+1}.
    \end{aligned}
\end{equation}
Taking logarithm on both sizes, we have
\begin{equation}
    \begin{aligned}
        \log\beta_t(s_t)&=\log\mathbb E_{s_{t+1}}\max_{a_t}\exp(\gamma^tr(s_t,a_t)+\log\beta_{t+1}(s_{t+1})).\\
    \end{aligned}
\end{equation}
Let $W(s_t)=\log \beta_t(s_t)$, we get
\begin{equation}
    W(s_t)=\log\mathbb E_{s_{t+1}}\exp\left[\max\limits_{a_t}\gamma^tr(s_t,a_t)+W(s_{t+1})\right].
\end{equation}
According to~\cite{levine2018reinforcement}, $W_t$ is a soft version of the traditional value function $V_t$. As the Soft Actor-Critic~\cite{haarnoja2018soft} has become the base algorithm in many scenarios, $\beta_t$ is closely related to the value function learned during training, which is often in its soft version. The forward probability $\alpha_t(s_t)=p\left(\mathcal{O}_{0: t-1} \mid s_t, \pi\right)$ is the probability of trajectory from timestep $0$ to $t-1$ being optimal given the state $s_t$. Such probability is hard to model as the transition from $s_{t-1}$ to $s_t$ is related to the actual policy $\pi$ as well as the environment dynamics. Therefore, we do not take $\alpha_t(s_t)$ into account when dividing training data to train the classifier.

\subsection{Discussions on the Surrogate for the Density Ratio}
\label{append:fgan}
According to some Off-Policy RL algorithms~\cite{liu2021regret, sinha2022experience}, the idea of training a classifier $D(s)$ as a data-based surrogate of the density ratio $\frac{\zeta(s)}{d_\pi(s)}$ can also be derived from a theorem related to f-divergence (lemma 1 in~\cite{sinha2022experience}).
Such derivation is essentially the same with our GAN-based proposition. Technically, these algorithms also divides the training data into two parts and train a classifier, which is later used to generate probabilities for prioritized sampling. Our SRPO algorithm proposes a different criterion to divide the training data, and train a classifier used in reward augmentation.
\subsection{Proofs to Theorems in Sec.~\ref{sec:theory}}
\label{append:proof}
We first introduce the following lemma which is essential in proving the two theorems in Sec.~\ref{sec:theory}.
\begin{lemma}
\label{th:v_diff}
Consider two homomorphous MDPs with dynamics $T$ and $T'$. Assuming $T'\in(T, \varepsilon_m)$, the reward function w.r.t. the action is $\lambda_1$-Lipschitz and the dynamics function w.r.t. the action is $\lambda_2$-inverse Lipschitz, we have 
\begin{equation}
    \left|V^*_T(s)-V^*_{T^\prime}(s)\right|\leqslant\frac{\lambda_1\lambda_2\varepsilon_m}{1-\gamma}
\end{equation}
for all $s\in\mathcal S$.
\end{lemma}
\begin{proof}
Recall the optimal value function under dynamics $T$ follows
\begin{equation}
    V^*_T(s)=\max\limits_a~r(s, a,T(s,a))+\gamma V^*_T(T(s,a)).
\end{equation}
Without the loss of generality, we assume $V^*_T(s)\geqslant V^*_{T^\prime}(s)$ on a certain state $s$. Define $a^*_T=\pi^*_T(s)$ and $\hat{a}$ such that $T'(s, \hat{a})=T(s, a^*_T)=s'$. Then we have
\begin{equation}
\label{eq:11}
\begin{aligned}
    \left|T(s,a^*_T)-T(s,\hat{a})\right|&=\left|T'(s,\hat{a})-T(s,\hat{a})\right|\leqslant\varepsilon_m,
\end{aligned}
\end{equation}
and
\begin{equation}
\label{eq:13}
    \left|r(s,a^*_T,s')-r(s,\hat{a},s')\right|\leqslant\lambda_1\left|a^*_T-\hat{a}\right|\leqslant\lambda_1\lambda_2\varepsilon_m.
\end{equation}
Therefore for all $s\in\mathcal S$,
\begin{equation}
    \begin{aligned}
    \left|V^*_T(s)-V^*_{T^\prime}(s)\right|&=V^*_T(s)-V^*_{T^\prime}(s)\\
    &=r(s,a^*_T,s')+\gamma V^*_T(s')-\max\limits_a~\left[r(s, a,T'(s,a))+\gamma V^*_{T'}(T'(s,a))\right] \\
    &\leqslant r(s,a^*_T,s')+\gamma V^*_T(s')-r(s, \hat{a},s')-\gamma V^*_{T'}(s') \\
    &\leqslant \lambda_1\lambda_2\varepsilon_m+ \gamma \left|V^*_T(s')-V^*_{T^\prime}(s')\right|\\
    &\leqslant \lambda_1\lambda_2\varepsilon_m+\gamma\lambda_1\lambda_2\varepsilon_m+ \gamma^2 \left|V^*_T(s'')-V^*_{T^\prime}(s'')\right|\\
    &\leqslant \cdots \\
    &\leqslant \frac{\lambda_1\lambda_2\varepsilon_m}{1-\gamma},
    \end{aligned}
\end{equation}
which concludes the proof.
\end{proof}

This lemma shows the discrepancy upper bound between the optimal state value functions in two homomorphous MDPs. We then apply it to prove the second theorem in Sec.~\ref{sec:theory}.
\begin{theorem}[Restatement of Thm.~\ref{cor:a_gap}]
Following the assumptions in Lem.~\ref{th:v_diff}, if the action gap $\Delta$ follows $\Delta>\frac{(2-\gamma)\lambda_1\lambda_2\varepsilon_m}{1-\gamma}$, for all $s\in\mathcal S$ we have $d^{*}_T(s)=d^{*}_{T'}(s)$.
\end{theorem}
\begin{proof}
Recall that the definition of action gap is $\Delta=\min\limits_{\theta\in\Theta}\min\limits_{s\in\mathcal{S}}\min\limits_{a\neq\pi_T^*(s)}V_{T_\theta}^*(s)-Q_{T_\theta}^*(s,a)$. Therefore, we have
\begin{equation}
\begin{aligned}
V^*_T(s)&\geqslant Q^*_T(s,a)+\Delta\\
&>Q^*_T(s,a)+\frac{(2-\gamma)\lambda_1\lambda_2\varepsilon_m}{1-\gamma}
\end{aligned}
\end{equation}
for all $(s,a)\in\mathcal S\times\mathcal A$ if $a\neq\pi^*_T(s)$. The same property holds for the transition function $T'$. We first show the state transition probability derived from $\pi^*_T$ and $\pi^*_{T'}$ is the same: $p_T(\cdot|s,\pi^*_T)=p_{T'}(\cdot|s,\pi^*_{T'}),~\forall s\in\mathcal S$. Without the loss of generality, let $V^*_T(s)\geqslant V_{T'}^*(s)(*)$. Let
\begin{equation}
    \begin{aligned}
        &\bar{a}=\argmax\limits_a r(s,a,T(s,a))+\gamma V_T^*(T(s,a))\\
        &a'=\argmax\limits_a r(s,a,T'(s,a))+\gamma V_{T'}^*(T'(s,a))\\
        &T'(s,\tilde{a})=T(s,\bar{a})=\bar{s},~T'(s,a')=s'.
    \end{aligned}
\end{equation}
According to Eq.~(\ref{eq:13}), $\|\tilde{a}-\bar{a}\|\leqslant\lambda_1\lambda_2\varepsilon_m$.
Supposing $\bar{s}\neq s'(**)$, we have $\tilde{a}\neq a'=\pi^*_{T'}(s)$. So
\begin{equation}
\label{eq:18-2}
    V^*_{T'}(s)>Q^*_{T'}(s,\tilde{a})+\frac{(2-\gamma)\lambda_1\lambda_2\varepsilon_m}{1-\gamma}.
\end{equation}
Meanwhile, 
\begin{equation}
\label{eq:19-2}
   \begin{aligned}
       Q^*_{T'}(s,\tilde{a})&=r(s,\tilde{a},\bar{s})+\gamma V^*_{T'}(\bar{s})\\
       &\geqslant r(s,\bar{a},\bar{s})+\gamma V^*_{T'}(\bar{s})-\lambda_1\lambda_2\varepsilon_m\\
       &\geqslant r(s,\bar{a},\bar{s})+\gamma V^*_{T}(\bar{s})-\lambda_1\lambda_2\varepsilon_m-\frac{\gamma\lambda_1\lambda_2\varepsilon_m}{1-\gamma}\\
       &=V^*_T(s)-\frac{(2-\gamma)\lambda_1\lambda_2\varepsilon_m}{1-\gamma}
   \end{aligned} 
\end{equation}
Combining Eq.~(\ref{eq:18-2}) and Eq.~(\ref{eq:19-2}), we get $V^*_{T'}(s)>V^*_T(s)$, which contradicts with Eq.~$(*)$. It means that the assumption $(**)$ is not correct, so $\bar{s}=s'$. 

We then show that $d^{*}_T(s)=d^{*}_{T'}(s)$ for all $s\in\mathcal S$:
\begin{equation}
    \begin{aligned}
    &\left\|p_T(s_t=\cdot|\pi_T^*)-p_{T'}(s_t=\cdot|\pi_{T'}^*)\right\|_\infty\\
    &=\left\|\sum_{s'}p_T(\cdot|s',\pi^*_T)p_T(s_{t-1}=s'|\pi_T^*)-p_{T'}(\cdot|s',\pi_{T'}^*)p_{T'}(s_{t-1}=s'|\pi_{T'}^*)\right\|_\infty\\
    &=\left\|\sum_{s'}p_T(\cdot|s',\pi^*_T)\left[p_T(s_{t-1}=s'|\pi_T^*)-p_{T'}(s_{t-1}=s'|\pi_{T'}^*)\right]\right\|_\infty\\
    &\leqslant\left\|\sum_{s'}p_T(\cdot|s',\pi^*_T)\left\|p_T(s_{t-1}=\cdot|\pi_T^*)-p_{T'}(s_{t-1}=\cdot|\pi_{T'}^*)\right\|_\infty\right\|_\infty\\
    &=\left\|\left\|p_T(s_{t-1}=\cdot|\pi_T^*)-p_{T'}(s_{t-1}=\cdot|\pi_{T'}^*)\right\|_\infty\sum_{s'}p_T(\cdot|s',\pi^*_T)\right\|_\infty\\
    &=\left\|p_T(s_{t-1}=\cdot|\pi_T^*)-p_{T'}(s_{t-1}=\cdot|\pi_{T'}^*)\right\|_\infty\\
    &\leqslant\cdots\\
    &\leqslant\left\|p_T(s_0=\cdot|\pi^*_T)-p_{T'}(s_0=\cdot|\pi_{T'}^*)\right\|_\infty\\
    &=0.
    \end{aligned}
\end{equation}
Therefore, for all $s\in\mathcal S$, we have $p_T(s_t=s|\pi_T^*)=p_{T'}(s_t=s|\pi_{T'}^*)$. So
\begin{equation}
    \left|d^{*}_T(s)-d^{*}_{T'}(s) \right|=\left|\sum_{t=0}^\infty p_T(s_t=s|\pi^*_T)-p_{T'}(s_t=s|\pi_{T'}^*)\right|=0
\end{equation}
for all $s\in\mathcal S$, which concludes the proof.
\end{proof}

Before proving the first theorem in Sec.~\ref{sec:theory}, we introduce a lemma that incorporates the 1-Wasserstein distance between the policies. It also considers a reference policy that has the same stationary state distribution with the optimal policy in the other dynamics. Such policy exists thanks to the homomorphous property of the MDPs.
\begin{lemma}
\label{lem:ori}
    Following the assumptions in Lem.~\ref{th:v_diff}, for all policy $\hat{\pi}$ such that
    $d^{\hat{\pi}}_T(s)=d^{*}_{T'}(s)$ for all $s\in\mathcal{S}$ and $\max _s W_1\left(\hat{\pi}(\cdot|s),\pi^*_{T'}(\cdot|s)\right) \leqslant \epsilon_\pi$, we have 
\begin{equation}
    \left|\eta_T(\pi_T^*)-\eta_T(\hat{\pi})\right|\leqslant\dfrac{\lambda_1\lambda_2\varepsilon_m+\lambda_1\varepsilon_\pi}{1-\gamma},
\end{equation}
where $W_1(\hat{\pi}(\cdot|s),\pi^*_{T'}(\cdot|s))$ is the 1-Wasserstein distance between two policies.
\end{lemma}


\begin{proof}
First, $|\eta_T(\pi^*_T)-\eta_{T'}(\pi^*_{T'})|$ can be bounded with Thm.~\ref{th:v_diff}:
\begin{equation}
\begin{aligned}
    \left|\eta_T(\pi^*_T)-\eta_{T'}(\pi^*_{T'})\right|&=\left|\mathbb E_{s\in\rho_0}V^*_T(s)-\mathbb{E}_{s\in\rho_0}V_{T'}^{*}(s)\right|\\
    &\leqslant\frac{\lambda_1\lambda_2\varepsilon_m}{1-\gamma}.
\end{aligned}
\end{equation}
We then try to bound $|\eta_{T}(\hat{\pi})-\eta_{T'}(\pi^*_{T'})|$. We first define the state-action stationary distributions $D_1(s,a)=d_T^{\hat{\pi}}(s)\hat{\pi}(a|s)$ and $D_2(s,a)=d^*_{T'}(s)\pi^*_{T'}(a|s)$. The accumulated return can be written as 
\begin{equation}
\label{eq:15}
    \begin{aligned}
        \eta_{T}(\hat{\pi})&=\frac{1}{1-\gamma}\mathbb E_{s,a,s'\sim D_1}\left[r(s,a,s')\right]\\
        \eta_{T'}(\pi^*_{T'})&=\frac{1}{1-\gamma}\mathbb E_{s,a,s'\sim D_2}\left[r(s,a,s')\right]\\
    \end{aligned}
\end{equation}

We start from the Lipschitz property of the reward function:
\begin{equation}
        |r(s,a_1,s')-r(s,a_2,s')|\leqslant \lambda_1\|a_1-a_2\|_1.
\end{equation}
Taking expectation w.r.t. $d^*_{T'}(\cdot)$ on both sides, we get
\begin{equation}
        \mathbb E_{s\sim d^*_{T'}}|r(s,a_1,s')-r(s,a_2,s')|\leqslant \mathbb E_{s\sim d^*_{T'}} \lambda_1\|a_1-a_2\|_1.
\end{equation}
Letting $\mu(A_1,A_2|s)$ be any joint distribution with marginals $\hat{\pi}$ and $\pi^*_{T'}$ conditioned on $s$. Taking expectation w.r.t. $\mu$ on both sides, we get
\begin{equation}
\label{eq:18}
\begin{aligned}
       \left|\mathbb E_{s,a\sim D_1}r(s,a,s')-\mathbb E_{s,a\sim D_2}r(s,a,s')\right|&\leqslant \mathbb E_{s\sim d^*_{T'}}\mathbb E_{a_1,a_2\sim\mu}|r(s,a_1,s')-r(s,a_2,s')|\\
       &\leqslant \lambda_1 \mathbb E_{s\sim d^*_{T'}}E_\mu\|a_1-a_2\|_1\\
       &\leqslant \max\limits_s\lambda_1 E_\mu\|a_1-a_2\|_1.
\end{aligned}
\end{equation}
Eq.~(\ref{eq:18}) holds for all joint distribution $\mu$, so it also holds for $\bar{\mu}=\argmin\limits_\mu\lambda_1E_\mu\|a_1-a_2\|_1$, leading to the 1-Wasserstein distance:
\begin{equation}
    \left|\mathbb E_{s,a\sim D_1}r(s,a,s')-\mathbb E_{s,a\sim D_2}r(s,a,s')\right| \leqslant\max_s\lambda_1W_1(\hat{\pi}(\cdot|s),\pi^*_{T'}(\cdot|s)).
\end{equation}
According to Eq.~(\ref{eq:15}), we have
\begin{equation}
    |\eta_{T}(\hat{\pi})-\eta_{T'}(\pi^*_{T'})|\leqslant\frac{\lambda_1\varepsilon_m}{1-\gamma}.
\end{equation}
Applying the triangle inequality, we get
\begin{equation}
    \begin{aligned}
         \left|\eta_T(\pi_T^*)-\eta_T(\hat{\pi})\right|&\leqslant|\eta_T(\pi^*_T)-\eta_{T'}(\pi^*_{T'})|+|\eta_{T}(\hat{\pi})-\eta_{T'}(\pi^*_{T'})|\\
         &\leqslant\dfrac{\lambda_1\lambda_2\varepsilon_m+\lambda_1\varepsilon_\pi}{1-\gamma},
    \end{aligned}
\end{equation}
which concludes the proof.

We then use this lemma to prove the first theory in Sec.~\ref{sec:theory}.
\end{proof}
\begin{theorem}[Restatement of Thm.~\ref{th:v_diff2}]
Consider two homomorphous MDPs with dynamics $T$ and $T'$. If $T'\in(T, \varepsilon_m)$, for all learning policy $\hat{\pi}$ such that $D_{\mathrm{KL}}(d^{\hat{\pi}}_T(\cdot)\|d^{*}_{T'}(\cdot))\leqslant\varepsilon_s$, we have
 \begin{equation}
\label{eq:return_diff_2}
    \eta_T(\hat{\pi})\geqslant\eta_T(\pi_T^*)-\dfrac{\lambda_1\lambda_2\varepsilon_m+2\lambda_1+\sqrt{2}R_{\max}\sqrt{\varepsilon_s}}{1-\gamma}.
\end{equation}
\end{theorem}
\begin{proof}
In two homomorphous MDPs with dynamics $T$ and $T'$, there exists a policy $\tilde\pi$ such that $d^{\tilde\pi}_T(\cdot)=d^*_{T'}(\cdot)$. According to Lem.~\ref{lem:ori}, we have
\begin{equation}
\label{eq:30}
    \begin{aligned}
            \left|\eta_T(\pi_T^*)-\eta_T(\tilde{\pi})\right|&\leqslant\dfrac{\lambda_1\lambda_2\varepsilon_m+\lambda_1\varepsilon_\pi}{1-\gamma}
            \leqslant\dfrac{\lambda_1\lambda_2\varepsilon_m+2\lambda_1}{1-\gamma},
    \end{aligned}
\end{equation}
where the second inequality is obtained as the actions are bounded to $[-1,1]$. The scaling is multiplied by the Lipschitz coefficient which tends to small, so it will make little influence to the bound. On the other hand, policies $\hat\pi$ and $\tilde\pi$ have a similar state discrepancy: $D_{\mathrm{KL}}(d^{\hat{\pi}}_T(\cdot)\|d^{\tilde\pi}_{T}(\cdot))\leqslant\varepsilon_s$. Therefore, their performance gap can be bounded according to results in imitation learning (Lem. 6 in~\cite{xu2020error}):
\begin{equation}
\label{eq:31}
    |\eta_T(\hat{\pi})-\eta_T(\tilde\pi)|\leqslant\frac{\sqrt{2}R_{\max}\sqrt{\varepsilon_s}}{1-\gamma}.
\end{equation}
Merging Eq.~(\ref{eq:30}) and Eq.~(\ref{eq:31}), we obtain
\begin{equation}
    \begin{aligned}
        \eta_T(\pi_T^*)-\eta_T(\hat\pi)&\leqslant|\eta_T(\hat\pi)-\eta_T(\pi_T^*)|\\
        &\leqslant\left|\eta_T(\pi_T^*)-\eta_T(\tilde{\pi})\right|+|\eta_T(\hat{\pi})-\eta_T(\tilde\pi)|\\
        &\leqslant\dfrac{\lambda_1\lambda_2\varepsilon_m+2\lambda_1+\sqrt{2}R_{\max}\sqrt{\varepsilon_s}}{1-\gamma}.
    \end{aligned}
\end{equation}
\end{proof}
\subsection{Discussions on the Theoretical Analysis}
\label{append:discussions}
\paragraph{The Lipschitz Assumptions in Sec.~\ref{sec:theory}}
Regarding the reward functions, the Lipschitz property implies that if $s$ and $s^{\prime}$ keeps unchanged, the deviation of the reward $r$ will be no larger than $\lambda_1$ times the deviation of the action $a$. Therefore, the Lipstchiz coefficient $\lambda_1$ is solely related to action-related terms in the reward function. It is important to note that different actions exist given the same $s$ and $s^{\prime}$ since we may compute the reward function in different dynamics. Considering the dynamics functions, the Lipschitz property indicates that if the current state $s$ remains unchanged and the actions differ by : $\left|a_1-a_2\right| \geqslant \varepsilon$, the next states will exhibit a significant difference: $\left|s_1^{\prime}-s_2^{\prime}\right| \geqslant \frac{\varepsilon}{\lambda 2}$.
\begin{table}[t]
    \centering
    \caption{$\lambda_1,\lambda_2,R_{\max}$ in practical environments.}
\begin{tabular}{lcccc} 
\toprule
Environment & Action-related Reward & $\lambda_1$ & $\lambda_2$ & $R_{\max}$ \\
\midrule CartPole-v0 & 0 & 0 & 1.42 & 1.00 \\
InvertedPendulum-v2 & 0 & 0 & 8.58 & 1.00 \\
Swimmer-v2 & $-0.0001|a|_2^2$ & 0.0001 & 2.59 &0.36 \\
HalfCheetah-v2 & $-0.1|a|_2^2$ & 0.1 & 1.01 &4.80 \\
Hopper-v2 & $-0.001|a|_2^2$ & 0.001 & 3.45& 3.80 \\
Walker2d-v2 & $-0.001|a|_2^2$ & 0.001 & 4.70 & $\geqslant4$ \\
Ant-v2 & $-0.5|a|_2^2$ & 0.5 & 0.69 & 6.0 \\
Humanoid-v2 & $-0.1|a|_2^2$ & 0.1 & 0.03 & $\geqslant8$ \\
\bottomrule
\end{tabular}
    \label{tab:append_lips}
\end{table}

In Tab.~\ref{tab:append_lips}, we list the action-related terms of the reward functions for various RL evaluation environments, along with the corresponding values of $\lambda_1$ derived from these terms. Additionally, we sample 50,000 $(s,a,s')$ tuples from the replay buffer, slightly modify the action, and observe how the resulting next state $s^{\prime}$ changes. The replay buffer contains trajectories collected during different training phases and should be diverse enough to cover most possible trajectories. This empirical analysis allows us to calculate $\lambda_2$ in practice. As indicated in the table, the action-related terms in reward functions exhibit reasonably small coefficients in all environments, leading to small $\lambda_1$ values. Combined with medium values of $\lambda_2$, it can be inferred that Lipschitz terms, including $\lambda_1$ and $\lambda_1 \lambda_2$, will remain small in practical scenarios and will not dominate the error term in Eq. 7. Also, the action gap assumption in Thm. 4.3 (line 258-259) is not strong and holds in many situations.
\paragraph{Failure Cases}
Although the assumptions are weak and hold in many situations, there are certain scenarios that these assumptions do not hold and the performance of SRPO can not be guaranteed. For example, in maze environments with different obstacle layout, the requirement of homomorphous MDPs is violated. There are also cases where the Lipstchitz coefficients $\lambda_1,\lambda_2$ can be large, such as stock markets with very high transaction feeds.

\paragraph{The Assumption on Dynamics Discrepancy}
We mentioned in Sec.~\ref{sec:theory} that one of the assumptions to prove the theorems is that $T\in(T',\varepsilon_m)$. In fact, this is a simplification of the actual requirement, which is weaker than the uniform bound of dynamics shift.  According to Eq.~(\ref{eq:11}), for any state $s$ we only require $T(s,a)$ and $T'(s,a)$ to be close on one specific action $\hat{a}$ such that $T^{\prime}(s, \hat{a})=T\left(s, a_T^*\right)=s^{\prime}$. This is a point-wise bound on dynamics shift and is comparable to assumptions in previous analysis~\cite{song2020provably}.
\paragraph{The Tightness of Eq.~(\ref{eq:return_diff})}
Eq.~(\ref{eq:return_diff}) has a similar form with the Eq.~(1) in Thm. 4.1 of~\cite{janner2019when}, where the return discrepancy $\left|\eta_T(\pi_T^*)-\eta_T(\hat{\pi})\right|$ is also bounded by differences in the policy distribution and transition functions, with an order of two in the effective horizon~(i.e. with a coefficient $\frac{1}{(1-\gamma)^2}$). By introducing some assumptions and constraining the policy to have the same stationary state distribution, we obtain a tighter discrepancy bound with an order of one in the effective horizon~(i.e. with a coefficient $\frac{1}{1-\gamma}$).

\section{Experiment Details}
\subsection{Setup}
\label{append:exp_setup}
To generate environments with different transition functions, We alter the xml file of the MuJoCo simulator and change its environment parameters. In online experiments, we build our code based on the Github repository\footnote{https://github.com/younggyoseo/CaDM/tree/master} of CaDM~\cite{lee2020context}. Some customized MuJoCo environments are defined in this repository. They have different reward functions with the original environments. We keep these modifications to make our online results comparable with the original CaDM algorithm. In offline experiments, we build our code based on the Github repository\footnote{https://github.com/polixir/OfflineRL}.
The offline datasets are generated by concatenating the data sampled in the original MuJoCo simulator, as well as simulators whose gravity and medium density are altered.  In both online and offline experiments, the evaluation is done in online static environments with all possible values of environment parameters. The average of these evaluation results is reported.
\begin{figure}[t]
    \centering
    \includegraphics[width=0.98\linewidth]{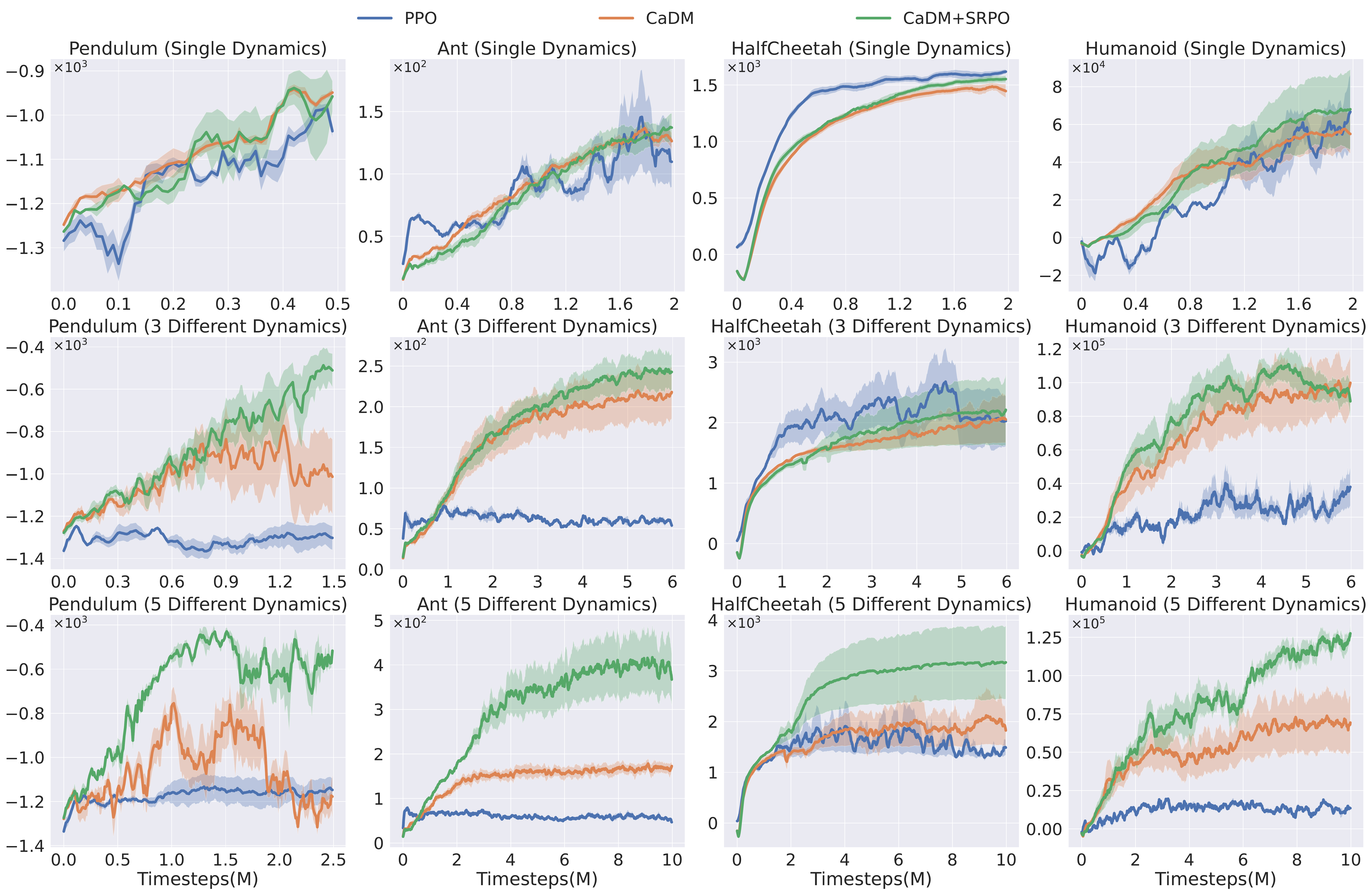}
    \caption{Detailed results of online experiments on MuJoCo tasks. In environments with single dynamics, three algorithms have a similar performance.}
    \label{fig:online_append}
\end{figure}
\input{offline_table_append}

\begin{table}[t]
\small
    \centering
    \caption{Results of offline experiments with a small dataset.}
    \label{tab:offline_small_dataset}
\begin{tabular}{l|c|c|c|c|} 
    \toprule
& MOPO & MAPLE & \makecell{MAPLE+\\DARA} & \makecell{MAPLE+\\SRPO (ours)} \\
\midrule Walker2d-medium & $0.21 \pm 0.13$ & $0.45 \pm 0.18$ & $0.74 \pm 0.12$ & $\mathbf{0 . 7 9} \pm 0.04$ \\
 Walker2d-medium-expert & $0.14 \pm 0.06$ & $0.26 \pm 0.01$ & $0.38 \pm 0.03$ & $\mathbf{0 . 6 1} \pm 0.11$ \\
 Hopper-medium & $0.01 \pm 0.00$ & $0.42 \pm 0.36$ & $0.36 \pm 0.06$ & $\mathbf{0 . 5 1} \pm 0.14$ \\
 Hopper-medium-expert & $0.01 \pm 0.00$ & $0.33 \pm 0.09$ & $0.16 \pm 0.04$ & $\mathbf{0 . 4 0} \pm 0.06$ \\
 HalfCheetah-medium & $0.10 \pm 0.01$ & $0.50 \pm 0.06$ & $0.37 \pm 0.01$ & $\mathbf{0 . 5 5} \pm 0.03$ \\
 HalfCheetah-medium-expert & $-0.03 \pm 0.00$ & $0.35 \pm 0.01$ & $\mathbf{0 . 6 3} \pm 0.03$ & $0.62 \pm 0.19$ \\
 \midrule
 Average & 0.07 & 0.39 & 0.44 & $\mathbf{0 . 5 8}$\\
 \bottomrule
\end{tabular}
\end{table}
\subsection{Additional Results}
\label{append:exp_res}
We show full results of online experiments on MuJoCo tasks in Fig.~\ref{fig:online_append}. Experiments on environments with single dynamics are included. These experiments are equivalent to those on static static environments. PPO, CaDM and CaDM+SRPO have a similar performance in these tasks. Full results of ablation studies are shown in Tab.~\ref{tab:ablation_append}. We also reduce the amount of offline data to 1/3 and perform additional experiments. The results are shown in Tab.~\ref{tab:offline_small_dataset}. MAPLE+SRPO can still achieve better performance than baseline algorithms. It improves the performance by 31\% over MAPLE+DARA and 49\% over MAPLE. These evidences indicate that SRPO indeed enables efficient data reuse, which is in accordance with statements in the introduction part.
\subsection{Additional Analysis}
\label{append:exp_analysis}

\begin{figure}{t}
    \centering
    \includegraphics[width=0.48\linewidth]{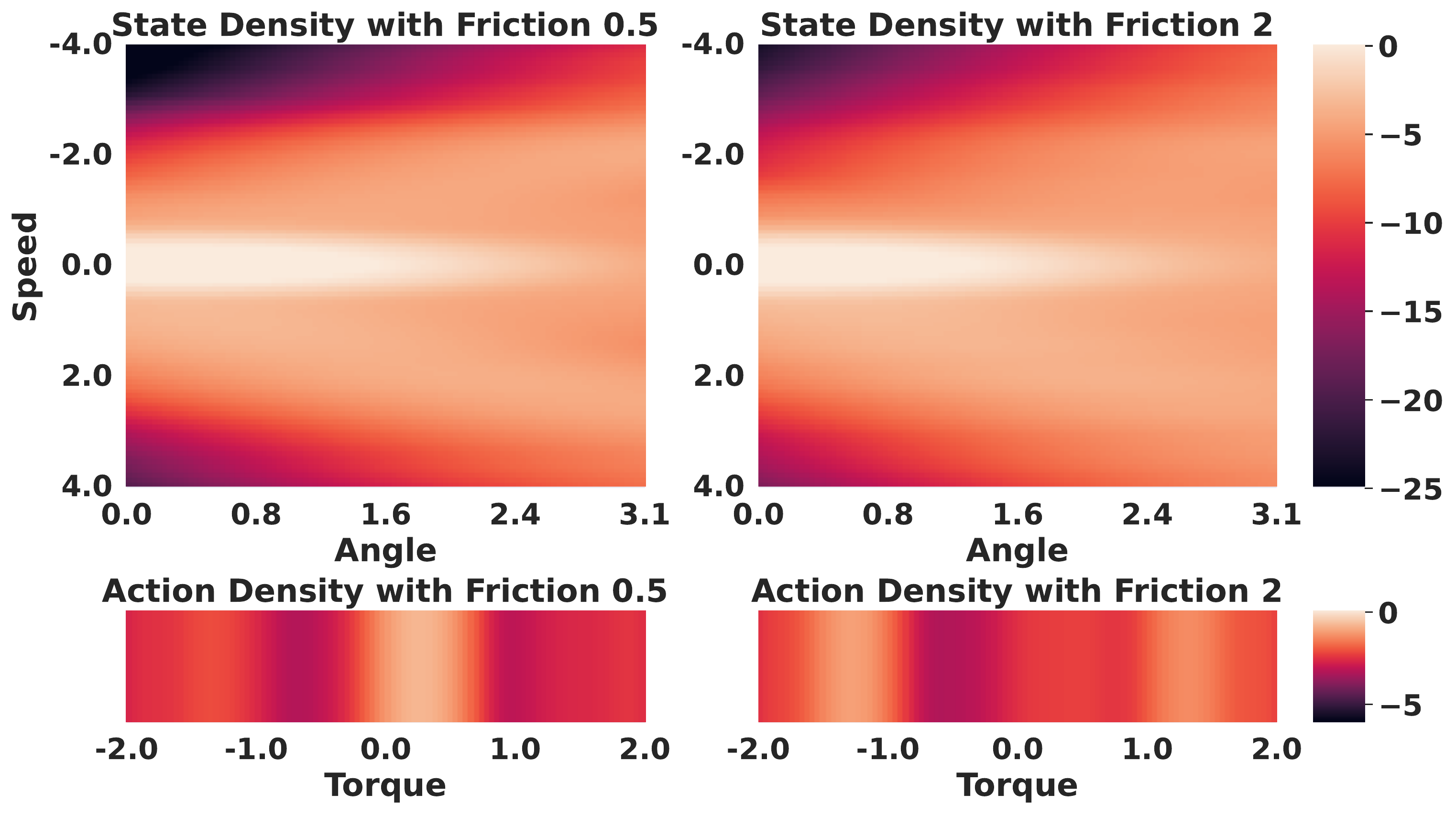}
    \caption{Visualization of state and action densities in data sampled from the Inverted Pendulum environment with 0.5 and 2 times of the original friction. 
    Under both frictions, the state distribution has high density with low pendulum speed and small pendulum angle. Meanwhile, the action distribution has different peaks in density under different frictions.}
    \label{fig:append_ex}
\end{figure}

To provide an additional demonstrating example to the intuition in Sec.~\ref{sec:motivating}, we train two policies in the Pendulum environment with 0.5 and 2 times of the original frictions and then visualize state and action densities. The results in Fig.~\ref{fig:append_ex} are similar to the experiments altering the environment gravity. We observe similar state distributions and different peaks in action distributions.

With respect to different Offline RL tasks, MAPLE+SRPO gains the highest rise in the Hopper environment and outperform all baseline methods in all of the 4 tasks. In the Walker2d and HalfCheetah environments, however, MAPLE+SRPO only outperforms in half of the tasks. Such difference results from the existence of multiple optimal policies which lead to different stationary state distributions~\cite{eysenbach2019diversity, zeng2022apd}. For example, the agent in the Walker2d environment has many ways to swing its arms to keep balance.  When the policy pattern in the offline dataset is different from the learning policy, its stationary state distribution may not be a good regularizer. The Hopper agent has a fewer degree of freedom compared with the other two, so the policy benefits more from regularizing with SRPO.

%% file: offline_table_append.tex
\newcommand{\rowll}[9]{#1 & #3 & #4 & #5 & #6 & #7 \\}
\newcommand{\rowss}[5]{#1 & #2 & #3 & #4 & #5 \\}
\begin{table}[t]
\small
    \centering
    \caption{Detailed results of ablation studies in offline experiments.}
    \label{tab:ablation_append}
    \begin{tabular}{l|c|c|c|c|c|c}
    \toprule
        ~ & \rowss{ \makecell{MAPLE\\+SRPO} }{\makecell{Behavior\\Regularization}}{\makecell{Random\\ Partition}}{\makecell{Fixed \\$\lambda=0.1$}}{\makecell{Fixed \\$\lambda=0.3$}}
        \midrule
\rowll{Walker2d-medium-expert    }{ 1.03$\pm$0.10 }{  0.66$\pm$0.08   }{\textbf{0.70}$\pm$0.18 }{0.42$\pm$0.16  }{0.66$\pm$0.08 }{0.38$\pm$0.16 }{0.80$\pm$0.02     }{1.11   }
\rowll{Walker2d-medium           }{0.78$\pm$0.01 }{\textbf{0.84}$\pm$0.03 }{0.71$\pm$0.02 }{0.79$\pm$0.00  }{0.72$\pm$0.13 }{\textbf{0.84}$\pm$0.03 }{0.83$\pm$0.03     }{0.79       }
\rowll{Walker2d-medium-replay    }{0.07$\pm$0.00 }{  \textbf{0.17}$\pm$0.02   }{0.16$\pm$0.01 }{0.14$\pm$0.01  }{\textbf{0.17}$\pm$0.02 }{0.16$\pm$0.01 }{0.17$\pm$0.01     }{0.27   }
\rowll{Walker2d-random           }{0.03$\pm$0.01 }{\textbf{0.22}$\pm$0.00 }{\textbf{0.22}$\pm$0.00 }{\textbf{0.22}$\pm$0.00  }{\textbf{0.22}$\pm$0.00 }{\textbf{0.22}$\pm$000  }{\textbf{0.22}$\pm$0.00 }{0.07       }
\rowll{Hopper-medium-expert      }{.32$\pm$0.14 }{\textbf{0.98}$\pm$0.02 }{0.85$\pm$0.25 }{0.46$\pm$0.14  }{\textbf{0.98}$\pm$0.02 }{0.86$\pm$0.18 }{0.96$\pm$0.06     }{0.98   }
\rowll{Hopper-medium             }{0.57$\pm$0.16 }{\textbf{1.03}$\pm$0.01 }{0.78$\pm$0.26 }{0.76$\pm$0.21  }{0.53$\pm$0.13 }{\textbf{1.03}$\pm$0.01 }{0.40$\pm$0.05     }{0.58       }
\rowll{Hopper-medium-replay      }{0.14$\pm$0.02 }{\textbf{1.02}$\pm$0.01 }{0.94$\pm$0.04 }{0.91$\pm$0.08  }{1.02$\pm$0.01 }{0.93$\pm$0.03 }{\textbf{1.02}$\pm$0.01 }{0.46       }
\rowll{Hopper-random             }{0.11$\pm$0.00 }{\textbf{0.32}$\pm$0.02 }{0.13$\pm$0.01 }{0.12$\pm$0.01  }{0.13$\pm$0.01 }{\textbf{0.32}$\pm$0.02 }{0.13$\pm$0.01     }{0.11       }
\rowll{Halfcheetah-medium-expert }{0.03$\pm$0.04 }{0.63$\pm$0.01 }{\textbf{0.65}$\pm$0.01 }{0.44$\pm$0.18  }{0.63$\pm$0.01 }{0.52$\pm$0.00 }{0.50$\pm$0.00     }{0.62       }
\rowll{Halfcheetah-medium        }{0.43$\pm$0.03 }{  \textbf{0.63}$\pm$0.01   }{0.60$\pm$0.00 }{0.62$\pm$0.02  }{0.61$\pm$0.02 }{\textbf{0.63}$\pm$0.01 }{0.67$\pm$0.03 }{0.44       }
\rowll{Halfcheetah-medium-replay }{0.46$\pm$0.00 }{\textbf{0.55}$\pm$0.00 }{0.54$\pm$0.00 }{0.54$\pm$0.01  }{\textbf{0.55}$\pm$0.00 }{0.24$\pm$0.01 }{0.53$\pm$0.01     }{0.46       }
\rowll{Halfcheetah-random        }{0.01$\pm$0.02 }{  \textbf{0.24}$\pm$0.01   }{0.21$\pm$0.03 }{0.20$\pm$0.01  }{\textbf{0.24}$\pm$0.01 }{0.23$\pm$0.01 }{0.21$\pm$0.00     }{0.35   }
\midrule
\rowll{Average                   }{    0.33      }{    \textbf{0.61}   }{0.54         }{0.47           }{0.54         }{0.53          }{0.54              }{0.52       }
        \bottomrule
    \end{tabular}
\end{table}


%% file: neurips2023.bbl
\begin{thebibliography}{10}

\bibitem{mnih2013playing}
Volodymyr Mnih, Koray Kavukcuoglu, David Silver, Alex Graves, Ioannis
  Antonoglou, Daan Wierstra, and Martin~A. Riedmiller.
\newblock Playing atari with deep reinforcement learning.
\newblock {\em CoRR}, abs/1312.5602, 2013.

\bibitem{silver2016mastering}
David Silver, Aja Huang, Chris~J. Maddison, Arthur Guez, Laurent Sifre, George
  van~den Driessche, Julian Schrittwieser, Ioannis Antonoglou, Vedavyas
  Panneershelvam, Marc Lanctot, Sander Dieleman, Dominik Grewe, John Nham, Nal
  Kalchbrenner, Ilya Sutskever, Timothy~P. Lillicrap, Madeleine Leach, Koray
  Kavukcuoglu, Thore Graepel, and Demis Hassabis.
\newblock Mastering the game of go with deep neural networks and tree search.
\newblock {\em Nat.}, 529(7587):484--489, 2016.

\bibitem{xue2022prefrec}
Wanqi Xue, Qingpeng Cai, Zhenghai Xue, Shuo Sun, Shuchang Liu, Dong Zheng, Peng
  Jiang, and Bo~An.
\newblock {PrefRec}: Preference-based recommender systems for reinforcing
  long-term user engagement.
\newblock {\em CoRR}, abs/2212.02779, 2022.

\bibitem{xue2023adarec}
Zhenghai Xue, Qingpeng Cai, Tianyou Zuo, Bin Yang, Lantao Hu, Peng Jiang, and
  Bo~An.
\newblock {AdaRec}: Adaptive sequential recommendation for reinforcing
  long-term user engagement.
\newblock {\em CoRR}, abs/2310.03984, 2023.

\bibitem{liu2022dara}
Jinxin Liu, Hongyin Zhang, and Donglin Wang.
\newblock {DARA:} {D}ynamics-aware reward augmentation in offline reinforcement
  learning.
\newblock In {\em {ICLR}}, 2022.

\bibitem{luo2022adapt}
Fan{-}Ming Luo, Shengyi Jiang, Yang Yu, Zongzhang Zhang, and Yi{-}Feng Zhang.
\newblock Adapt to environment sudden changes by learning a context sensitive
  policy.
\newblock In {\em {AAAI}}, 2022.

\bibitem{schulman2017proximal}
John Schulman, Filip Wolski, Prafulla Dhariwal, Alec Radford, and Oleg Klimov.
\newblock Proximal policy optimization algorithms.
\newblock {\em CoRR}, abs/1707.06347, 2017.

\bibitem{haarnoja2018soft}
Tuomas Haarnoja, Aurick Zhou, Pieter Abbeel, and Sergey Levine.
\newblock Soft actor-critic: Off-policy maximum entropy deep reinforcement
  learning with a stochastic actor.
\newblock In {\em {ICML}}, 2018.

\bibitem{lee2020context}
Kimin Lee, Younggyo Seo, Seunghyun Lee, Honglak Lee, and Jinwoo Shin.
\newblock Context-aware dynamics model for generalization in model-based
  reinforcement learning.
\newblock In {\em {ICML}}, 2020.

\bibitem{chen2022latent}
Xi~Chen, Ali Ghadirzadeh, Tianhe Yu, Yuan Gao, Jianhao Wang, Wenzhe Li, Bin
  Liang, Chelsea Finn, and Chongjie Zhang.
\newblock Latent-variable advantage-weighted policy optimization for offline
  {RL}.
\newblock {\em CoRR}, abs/2203.08949, 2022.

\bibitem{zhou2019environment}
Wenxuan Zhou, Lerrel Pinto, and Abhinav Gupta.
\newblock Environment probing interaction policies.
\newblock In {\em {ICLR}}, 2019.

\bibitem{chen2021offline}
Xiong{-}Hui Chen, Yang Yu, Qingyang Li, Fan{-}Ming Luo, Zhiwei~(Tony) Qin,
  Wenjie Shang, and Jieping Ye.
\newblock Offline model-based adaptable policy learning.
\newblock In {\em NeurIPS}, 2021.

\bibitem{eysenbach2021off}
Benjamin Eysenbach, Shreyas Chaudhari, Swapnil Asawa, Sergey Levine, and Ruslan
  Salakhutdinov.
\newblock Off-dynamics reinforcement learning: Training for transfer with
  domain classifiers.
\newblock In {\em {ICLR}}, 2021.

\bibitem{niu2022when}
Haoyi Niu, Shubham Sharma, Yiwen Qiu, Ming Li, Guyue Zhou, Jianming Hu, and
  Xianyuan Zhan.
\newblock When to trust your simulator: Dynamics-aware hybrid
  offline-and-online reinforcement learning.
\newblock {\em CoRR}, abs/2206.13464, 2022.

\bibitem{joshua2017constrained}
Joshua Achiam, David Held, Aviv Tamar, and Pieter Abbeel.
\newblock Constrained policy optimization.
\newblock In {\em {ICML}}, 2017.

\bibitem{levine2018reinforcement}
Sergey Levine.
\newblock Reinforcement learning and control as probabilistic inference:
  Tutorial and review.
\newblock {\em CoRR}, abs/1805.00909, 2018.

\bibitem{velez2016hidden}
Finale Doshi{-}Velez and George~Dimitri Konidaris.
\newblock Hidden parameter markov decision processes: {A} semiparametric
  regression approach for discovering latent task parametrizations.
\newblock In {\em {IJCAI}}, 2016.

\bibitem{kirk2023survey}
Robert Kirk, Amy Zhang, Edward Grefenstette, and Tim Rockt{\"{a}}schel.
\newblock A survey of zero-shot generalisation in deep reinforcement learning.
\newblock {\em J. Artif. Intell. Res.}, 76:201--264, 2023.

\bibitem{luisa2020varibad}
Luisa~M. Zintgraf, Kyriacos Shiarlis, Maximilian Igl, Sebastian Schulze, Yarin
  Gal, Katja Hofmann, and Shimon Whiteson.
\newblock {VariBAD}: {A} very good method for bayes-adaptive deep {RL} via
  meta-learning.
\newblock In {\em {ICLR}}, 2020.

\bibitem{yang2020single}
Jiachen Yang, Brenden~K. Petersen, Hongyuan Zha, and Daniel~M. Faissol.
\newblock Single episode policy transfer in reinforcement learning.
\newblock In {\em {ICLR}}, 2020.

\bibitem{finn2017model}
Chelsea Finn, Pieter Abbeel, and Sergey Levine.
\newblock Model-agnostic meta-learning for fast adaptation of deep networks.
\newblock In {\em {ICML}}, 2017.

\bibitem{mitchell2021offline}
Eric Mitchell, Rafael Rafailov, Xue~Bin Peng, Sergey Levine, and Chelsea Finn.
\newblock Offline meta-reinforcement learning with advantage weighting.
\newblock In {\em {ICML}}, 2021.

\bibitem{tuomas2017reinforcement}
Tuomas Haarnoja, Haoran Tang, Pieter Abbeel, and Sergey Levine.
\newblock Reinforcement learning with deep energy-based policies.
\newblock In {\em {ICML}}, 2017.

\bibitem{pang2021sparsity}
Jing{-}Cheng Pang, Tian Xu, Shengyi Jiang, Yu{-}Ren Liu, and Yang Yu.
\newblock Sparsity prior regularized q-learning for sparse action tasks.
\newblock {\em CoRR}, abs/2105.08666, 2021.

\bibitem{fujimoto2019off}
Scott Fujimoto, David Meger, and Doina Precup.
\newblock Off-policy deep reinforcement learning without exploration.
\newblock In {\em {ICML}}, 2019.

\bibitem{kumar2020conservative}
Aviral Kumar, Aurick Zhou, George Tucker, and Sergey Levine.
\newblock Conservative q-learning for offline reinforcement learning.
\newblock In {\em NeurIPS}, 2020.

\bibitem{wu2019behavior}
Yifan Wu, George Tucker, and Ofir Nachum.
\newblock Behavior regularized offline reinforcement learning.
\newblock {\em CoRR}, abs/1911.11361, 2019.

\bibitem{islam2019off}
Riashat Islam, Komal~K. Teru, and Deepak Sharma.
\newblock Off-policy policy gradient algorithms by constraining the state
  distribution shift.
\newblock {\em CoRR}, abs/1911.06970, 2019.

\bibitem{liu2018breaking}
Qiang Liu, Lihong Li, and Ziyang~Tang and´ Dengyong~Zhou.
\newblock Breaking the curse of horizon: Infinite-horizon off-policy
  estimation.
\newblock In {\em NeurIPS}, 2018.

\bibitem{nachum2019dualdice}
Ofir Nachum, Yinlam Chow, Bo~Dai, and Lihong Li.
\newblock Dualdice: Behavior-agnostic estimation of discounted stationary
  distribution corrections.
\newblock In {\em NeurIPS}, 2019.

\bibitem{gangwni2020state}
Tanmay Gangwani and Jian Peng.
\newblock State-only imitation with transition dynamics mismatch.
\newblock In {\em {ICLR}}, 2020.

\bibitem{liu2020state}
Fangchen Liu, Zhan Ling, Tongzhou Mu, and Hao Su.
\newblock State alignment-based imitation learning.
\newblock In {\em {ICLR}}, 2020.

\bibitem{ni2019firl}
Tianwei Ni, Harshit~S. Sikchi, Yufei Wang, Tejus Gupta, Lisa Lee, and Ben
  Eysenbach.
\newblock {f-IRL}: Inverse reinforcement learning via state marginal matching.
\newblock In {\em CoRL}, 2020.

\bibitem{yang2022regularizing}
Shentao Yang, Yihao Feng, Shujian Zhang, and Mingyuan Zhou.
\newblock Regularizing a model-based policy stationary distribution to
  stabilize offline reinforcement learning.
\newblock In {\em {ICML}}, 2022.

\bibitem{christiano2016transfer}
Paul~F. Christiano, Zain Shah, Igor Mordatch, Jonas Schneider, Trevor
  Blackwell, Joshua Tobin, Pieter Abbeel, and Wojciech Zaremba.
\newblock Transfer from simulation to real world through learning deep inverse
  dynamics model.
\newblock {\em CoRR}, abs/1610.03518, 2016.

\bibitem{jiang2020offline}
Shengyi Jiang, Jing{-}Cheng Pang, and Yang Yu.
\newblock Offline imitation learning with a misspecified simulator.
\newblock In {\em NeurIPS}, 2020.

\bibitem{parzen1962kde}
Emanuel Parzen.
\newblock On estimation of a probability density function and mode.
\newblock {\em The annals of mathematical statistics}, 33(3):1065--1076, 1962.

\bibitem{goodfellow2014generative}
Ian~J. Goodfellow, Jean Pouget{-}Abadie, Mehdi Mirza, Bing Xu, David
  Warde{-}Farley, Sherjil Ozair, Aaron~C. Courville, and Yoshua Bengio.
\newblock Generative adversarial networks.
\newblock {\em CoRR}, abs/1406.2661, 2014.

\bibitem{nowozin2016fgan}
Sebastian Nowozin, Botond Cseke, and Ryota Tomioka.
\newblock f-gan: Training generative neural samplers using variational
  divergence minimization.
\newblock In {\em {NIPS}}, 2016.

\bibitem{yu2020mopo}
Tianhe Yu, Garrett Thomas, Lantao Yu, Stefano Ermon, James~Y. Zou, Sergey
  Levine, Chelsea Finn, and Tengyu Ma.
\newblock {MOPO:} model-based offline policy optimization.
\newblock In {\em NeurIPS}, 2020.

\bibitem{todorov2012mujoco}
Emanuel Todorov, Tom Erez, and Yuval Tassa.
\newblock Mujoco: {A} physics engine for model-based control.
\newblock In {\em {IROS}}, 2012.

\bibitem{fu2020d4rl}
Justin Fu, Aviral Kumar, Ofir Nachum, George Tucker, and Sergey Levine.
\newblock {D4RL:} datasets for deep data-driven reinforcement learning.
\newblock {\em CoRR}, abs/2004.07219, 2020.

\bibitem{jonothan2016generative}
Jonathan Ho and Stefano Ermon.
\newblock Generative adversarial imitation learning.
\newblock In {\em {NIPS}}, 2016.

\bibitem{kate2019efficient}
Kate Rakelly, Aurick Zhou, Chelsea Finn, Sergey Levine, and Deirdre Quillen.
\newblock Efficient off-policy meta-reinforcement learning via probabilistic
  context variables.
\newblock In {\em {ICML}}, 2019.

\bibitem{liu2021regret}
Xu{-}Hui Liu, Zhenghai Xue, Jing{-}Cheng Pang, Shengyi Jiang, Feng Xu, and Yang
  Yu.
\newblock Regret minimization experience replay in off-policy reinforcement
  learning.
\newblock In {\em NeurIPS}, 2021.

\bibitem{sinha2022experience}
Samarth Sinha, Jiaming Song, Animesh Garg, and Stefano Ermon.
\newblock Experience replay with likelihood-free importance weights.
\newblock In {\em {L4DC}}, 2022.

\bibitem{xu2020error}
Tian Xu, Ziniu Li, and Yang Yu.
\newblock Error bounds of imitating policies and environments.
\newblock In {\em NeurIPS}, 2020.

\bibitem{song2020provably}
Yuda Song, Aditi Mavalankar, Wen Sun, and Sicun Gao.
\newblock Provably efficient model-based policy adaptation.
\newblock In {\em {ICML}}, 2020.

\bibitem{janner2019when}
Michael Janner, Justin Fu, Marvin Zhang, and Sergey Levine.
\newblock When to trust your model: Model-based policy optimization.
\newblock In {\em NeurIPS}, 2019.

\bibitem{eysenbach2019diversity}
Benjamin Eysenbach, Abhishek Gupta, Julian Ibarz, and Sergey Levine.
\newblock Diversity is all you need: Learning skills without a reward function.
\newblock In {\em {ICLR}}, 2019.

\bibitem{zeng2022apd}
Kailin Zeng, Qiyuan Zhang, Bin Chen, Bin Liang, and Jun Yang.
\newblock {APD:} learning diverse behaviors for reinforcement learning through
  unsupervised active pre-training.
\newblock {\em {IEEE} Robotics Autom. Lett.}, 7(4):12251--12258, 2022.

\end{thebibliography}
